\documentclass[preprint,12pt]{elsarticle}

\usepackage{tikz}
\usetikzlibrary{shapes.geometric, arrows.meta, calc, positioning, decorations.pathreplacing}

\usepackage{amssymb}
\usepackage{amsthm,url,graphicx,booktabs,lipsum,color,bm,caption,subcaption,soul}

\usepackage{amsmath}
\usepackage{tabularx}
\usepackage{setspace}
\usepackage{svg}
\usepackage[gen]{eurosym}
\usepackage{float}
\usepackage{array}
\usepackage{tabularx}
\usepackage{multirow}
\usepackage{siunitx} 
\usepackage[table,xcdraw]{xcolor}

\usepackage[xcolor]{changebar}
\usepackage{lipsum}

\newcommand{\rev}[1]{\textcolor{black}{#1}}
\usepackage{natbib}

\usepackage{multirow}

\journal{Engineering Applications of Artificial Intelligence}

\begin{document}

\begin{frontmatter}



\title{WaveletInception Networks for on-board Vibration-Based Infrastructure Health Monitoring}

%

\author[label1]{Reza Riahi Samani}
\author[label2]{Alfredo Núñez}
\author[label1]{Bart De Schutter}

\affiliation[label1]{organization={Delft Center for Systems and Control (DCSC), Delft University of Technology},
            addressline={Mekelweg 2},
            postcode={2628 CD},
            city={Delft},
            country={The Netherlands}}

\affiliation[label2]{organization={Section of Railway Engineering, Department of Engineering Structures, Delft University of Technology},
            addressline={Stevinweg 1}, 
            postcode={2628 CN},
            city={Delft},
            country={The Netherlands}}


\begin{abstract}

\rev{This paper presents a deep learning framework for analyzing on-board vibration response signals in infrastructure health monitoring. The proposed WaveletInception–BiGRU network uses a Learnable Wavelet Packet Transform (LWPT) for early spectral feature extraction, followed by one-dimensional Inception-Residual Network (1D Inception-ResNet) modules for multi-scale, high-level feature learning. Bidirectional Gated Recurrent Unit (BiGRU) modules then integrate temporal dependencies and incorporate operational conditions, such as the measurement speed. This approach enables effective analysis of vibration signals recorded at varying speeds, eliminating the need for explicit signal preprocessing. The sequential estimation head further leverages bidirectional temporal information to produce an accurate, localized assessment of infrastructure health. Ultimately, the framework generates high-resolution health profiles spatially mapped to the physical layout of the infrastructure. Case studies involving track stiffness regression and transition zone classification using real-world measurements demonstrate that the proposed framework significantly outperforms state-of-the-art methods, underscoring its potential for accurate, localized, and automated on-board infrastructure health monitoring.}
\end{abstract}

\begin{graphicalabstract}
\begin{figure}[H]
      \centering
       \includegraphics[scale=0.65, trim=0 150 0 290,clip]{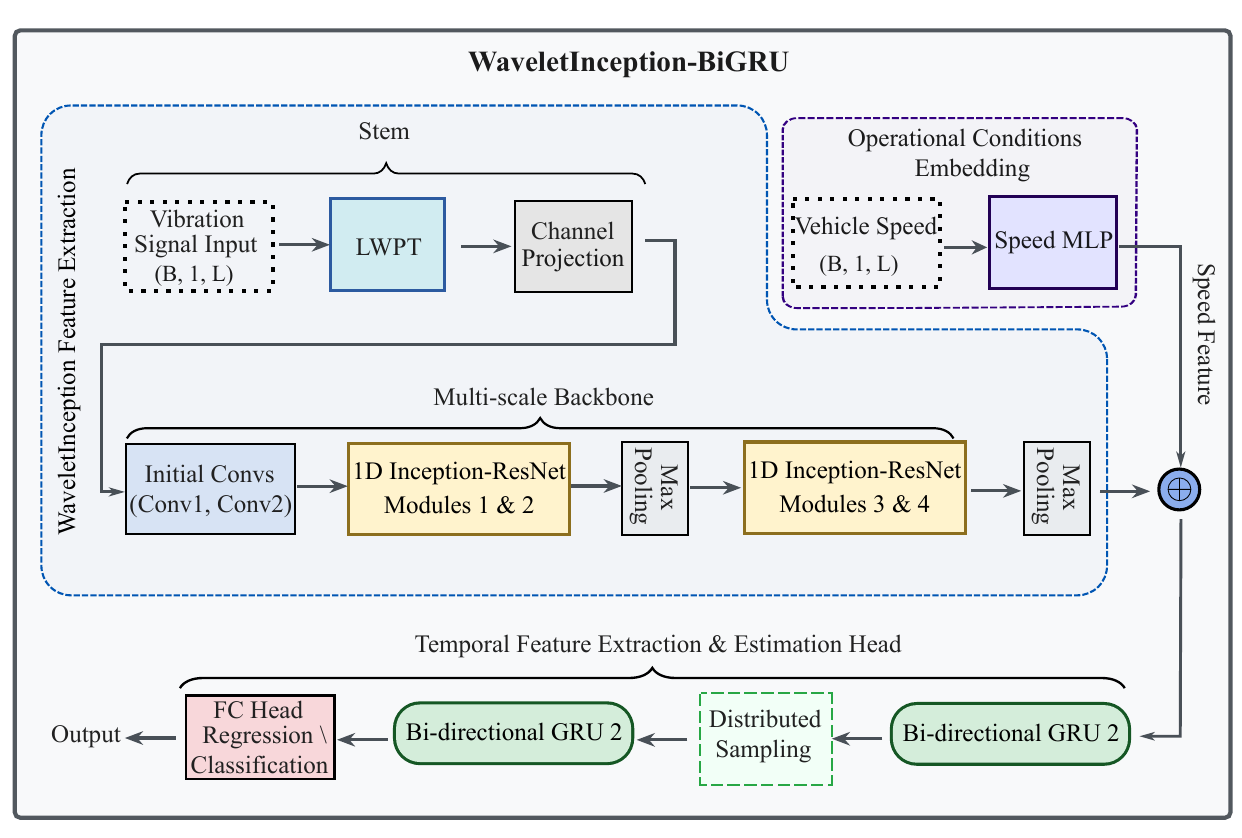}  
      \caption{Overview of the WaveletInception-BiGRU architecture, including WaveletInception vibration signal feature extraction, operational conditions feature fusion, and BiGRU sequential health conditions estimation.}
      \label{fg:method}
   \end{figure}
\end{graphicalabstract}

\begin{highlights}

\item \rev{WaveletInception-BiGRU is proposed for on-board monitoring of railway infrastructure.}
\item WaveletInception enables rich spectral and multi-scale feature extraction from on-board vibration signals.

\item \rev{The BiGRU-based network captures bidirectional temporal dependencies, and enables sequential modeling for localized health condition estimation.}

\item \rev{Late-stage BiGRU-based fusion automates the integration of operational conditions and vibration features, eliminating the need for manual feature engineering.}

\end{highlights}


\end{frontmatter}

\section{Introduction}

Efficient maintenance is fundamental to the sustainability, safety, and cost-effectiveness of transportation infrastructure. In 2020, for example, maintenance and renewal for railway infrastructure in Europe amounted to \euro10.47~billion and \euro11.24~billion, respectively, accounting for 52\% of total infrastructure-related expenditures \cite{REPORTCOMMISSIONEUROPEAN2023}. An effective maintenance strategy heavily relies on its underlying monitoring techniques. Early and accurate fault detection is crucial for minimizing downtime, extending service life, and reducing overall life cycle and maintenance costs. Traditionally, railway infrastructure health monitoring has relied on expert visual inspections or deploying sensing devices in strategic locations. However, these methods are often time-consuming, labor-intensive, and limited in scale. By overcoming such challenges, vehicle-based monitoring tools have gained significant attention  \cite{fernandez-bobadillaModernTendenciesVehicleBased2023}. On-board vibration response analysis is a powerful and widely adopted vehicle-based monitoring technique that enables cost-effective and frequent health assessments across the entire infrastructure network under operational conditions. In particular, in the context of railway track and bridge infrastructures, vibration signals collected by accelerometers mounted on operational vehicles offer a cost-effective and scalable means of assessing infrastructure conditions \cite{Sansiñena26032025, 11021483}. 

Several approaches have been reported in the literature for the analysis of on-board vibration response signals \cite{phusakulkajornArtificialIntelligenceRailway2023a}. At the core of this analysis lies feature extraction, a crucial step that involves identifying damage-sensitive features that can reflect the health status of the infrastructure. These features may be derived from the time and frequency domain \cite{ledermanTrackmonitoring2017}, time-frequency domain \cite{shenEvaluatingRailwayTrack2023a, lamprea-pinedaRailwayTrackReceptance2024}, and space-frequency domain \cite{unsiwilaiMultipleaxle2023}, with the latter often utilizing wavelet transforms to capture spectral information. Once extracted, these features are commonly fed into machine learning classifiers for classifying vibration signals or employed for developing track quality indices \cite{shenEvaluatingRailwayTrack2023a, unsiwilaiMultipleaxle2023}. While these approaches offer physical interpretations, they are often limited by the complexity of selecting the most relevant features and requiring experts' experience in defining handcrafted features, which can hinder an automated process with high performance.

In recent years, deep learning methods for analyzing on-board vibration responses have gained significant attention. However, they still encounter key challenges. Most notably, existing approaches often struggle to effectively capture the multi-scale, non-stationary characteristics of on-board vibration signals. Wavelet-inspired neural networks can decompose signals into time-frequency components, presenting a promising solution for extracting physically meaningful features from this data. Nonetheless, there is currently no fully automated, supervised learning framework that integrates these capabilities.  Moreover, existing deep learning frameworks overlook the incorporation of operational factors, such as measurement speed, rail profile, or track alignment, when analyzing on-board vibration data, failing to account for the influence of these factors. Specifically, current methods typically have difficulty accommodating varying measurement speeds, which limits their generalization in real-world applications under diverse operational conditions. To address these research gaps, this paper introduces a novel deep learning methodology for monitoring the health of railway infrastructure using on-board vibration responses. The key contributions of this work include:

\begin{itemize}
    \item \rev{A WaveletInception network is proposed for extracting features from vibration signals. It combines a learnable wavelet transform with one-dimensional Inception-Residual Network (1D Inception-ResNet) blocks to capture multi-scale information. The stem module directly incorporates spectral information, allowing the model to learn meaningful kernels in the early stages of learning. In the deeper layers, 1D Inception-ResNet blocks extract more nuanced features. This approach improves both accuracy and computational efficiency compared to existing feature extraction methods.}

    \item \rev{A feature fusion strategy is proposed to integrate operational conditions, such as measurement speed, rail profile, and track alignment, into vibration signal analysis. These conditions strongly influence excitation levels, signal length, and spatial frequency content. To explicitly model their interaction with learned features, a bidirectional Gated Recurrent Unit (BiGRU) module is introduced within the feature extraction network, enabling mid-level fusion of operational information and bidirectional temporal dependencies, thereby enabling analysis under different operational conditions.}

    \item \rev{The model processes time-domain signals of varying speeds and lengths without preprocessing, making it broadly applicable and suitable for automated monitoring.}

    \item \rev{A bidirectional sequential modeling strategy is employed in the estimation stage to exploit contextual information from neighboring signal segments. By leveraging bidirectional temporal dependencies in on-board vibration responses, the proposed framework achieves more accurate infrastructure condition estimation and finer localized assessment compared with unidirectional recurrent architectures.}

    \item \rev{The WaveletInception-BiGRU network improves estimation accuracy and computational cost over existing deep learning models for on-board vibration analysis.}

\end{itemize}

\rev{The remainder of this paper is organized as follows. Section \ref{related} reviews related work on on-board vibration signal analysis and recent deep learning advances. Section \ref{preli} presents the preliminaries. Section \ref{Method} describes the proposed model architecture. Section \ref{case} evaluates the model using two case studies, including one based on real-world railway measurements. Sections \ref{Disc} and \ref{Conc} discuss the results, summarize key findings, and outline future research directions.}

\section{Related work}\label{related}


Infrastructure health monitoring using on-board vibration response requires methodologies for extracting meaningful information from the vibration signals. Modal parameters, such as natural frequencies, mode shapes, and damping ratios, are widely recognized as damage-sensitive features \cite{yangStateoftheArt2018}. In addition, stiffness is often estimated for structural health monitoring purposes. Papers like \cite{quirkeDrivebyDetectionRailway2017, zhuIdentificationRailwayBallasted2018} have used optimization techniques, including cross-entropy and adaptive regularization, to estimate railway track stiffness by matching vehicle-track interaction models to vibration data.  However, these inverse identification methods are computationally intensive. For instance, estimating track stiffness for a 140-meter segment can take up to 11 hours \cite{quirkeDrivebyDetectionRailway2017}, which makes them impractical for (pseudo) real-time monitoring. Although advancements in computational power have reduced simulation times, solving inverse problems in railway infrastructure remains time-consuming due to the numerous possible parameter combinations, structural configurations, operational scenarios, and the distributed nature of railway tracks. This complexity significantly limits the feasibility of using inverse methods for (pseudo) real-time monitoring.

\rev{Other approaches have focused on extracting informative features from vibration signals for infrastructure monitoring. These features, often compact representations of the original signals, are mostly spectral features derived from Fourier transforms \cite{caprioliRail2007}, wavelet transforms \cite{shenFastRobustIdentification2021, shenEvaluatingRailwayTrack2023a, unsiwilaiMultipleaxle2023}, Kalman filters \cite{ hoelzlVoldKalmanFilter2023}, or Hilbert transforms \cite{malekjafarianRailway2019}, which are sensitive to structural health conditions. For example, features can be extracted in the time domain \cite{MAO2025115988}, time-frequency \cite{shenEvaluatingRailwayTrack2023a}, or space-frequency domain \cite{unsiwilaiMultipleaxle2023} by employing synchro-squeezed wavelet transforms, and continuous wavelets transform to construct localized wavelet power spectra. These approaches leverage the strength of wavelets to localize features along the signal. Moreover, statistical features derived from the power spectrum in the space domain are also robust to spatial frequency variations, making them practical for signals obtained at diverse measurement speeds. These features are used as health condition indicators, such as track quality indices \cite{unsiwilaiMultipleaxle2023}, or as inputs to machine learning models like Gaussian process regression \cite{shenEvaluatingRailwayTrack2023a} for estimating stiffness parameters. Papers like \cite{zhangDatadrivenWindinducedResponse2025, zhangBayesianDynamicModelling2024, zhangBayesianDynamicRegression2022, pengInterpretableDamageSensitive2025, khodadoostIntelligentVibrationbasedStructural2025} further discuss the challenges and opportunities of data-driven and machine learning approaches in vibration-signal-based infrastructure monitoring.}

\rev{While frameworks based on feature engineering effectively capture the spectral characteristics of signals, they heavily rely on carefully selected hyperparameters, such as window type, length, stride, and transformations (e.g., using log-mel spectrograms, which apply a logarithmic scale to Mel-frequency features). This reliance can pose challenges, as these hyperparameters significantly impact analysis accuracy, complicate full automation, and require time-consuming tuning \cite{gaoEfficientMultiScaleNetwork2024, zhangBayesianDynamic2022a}. Additionally, the frequency components extracted do not directly provide physical insights and typically necessitate further modeling or expert interpretation to connect them to specific structural behaviors \cite{unsiwilaiEnhanced2024, phusakulkajornArtificialIntelligenceRailway2023a}. However, supervised deep learning methods allow models to learn patterns from labeled data by minimizing prediction errors, leading to superior performance. These models can directly learn features and generalize more effectively than fixed wavelet coefficients, without the need for extensive fine-tuning and manual feature engineering.}

Recent studies have increasingly investigated the use of supervised deep learning methodologies for monitoring the health of infrastructure through on-board vibration responses. These frameworks have been applied to various tasks related to infrastructure monitoring, including track stiffness estimation \cite{shenEvaluatingRailwayTrack2023a, shenFastRobustIdentification2021, riahisamaniBidirectionalLongShortTerm2025, huangQuantification2022}, as instantaneous decreases in moment of inertia for bridge elements \cite{lockeUsing2020}, model bridge’s dynamic changes due to additional weight \cite{hajializadehDeep2023}, identification of seized bearings and cracking in the bridge beams \cite{corballyDeeplearning2024}, and detecting transition zone characteristics \cite{phusakulkajornHybridNeuralModel2025a}. Due to the high performance of deep learning methods in terms of computational efficiency and prediction accuracy, the literature indicates a growing application of these techniques for on-board vibration-based monitoring \cite{phusakulkajornArtificialIntelligenceRailway2023a, zhangDatadrivenWindinducedResponse2025, hajializadehDeep2023}.

Deep learning frameworks for the on-board vibration signal analysis are typically CNN-based \cite{huangQuantification2022, lockeUsing2020, hajializadehDeep2023, corballyDeeplearning2024, 10967036} or LSTM-based and hybrid architectures \cite{ phusakulkajornHybridNeuralModel2025a, samaniBidirectional2024}. For instance, \cite{lockeUsing2020} utilized a multi-layer CNN to classify bridge damage from simulated vehicle vibration signals. The model input includes the frequency response spectrum, adjusted to a spectral resolution of 0.1 Hz, combined with operational conditions like ambient temperature, speed, and vehicle mass.  \cite{lockeUsing2020} also demonstrated the robustness of CNNs to environmental and operational noise. \cite{huangQuantification2022} employed a dilated CNN to analyze simulated axle-box acceleration (ABA) signals for predicting railway track dynamic stiffness. By training models with three input window sizes in the time domain and with measurement speed variations (60–120 km/h), the approach estimates track stiffness over different infrastructure lengths. The dilated CNN required much less computation time compared to standard CNNs without any significant enhanced accuracy \cite{huangQuantification2022}. Transfer learning was explored in \cite{hajializadehDeep2023}, which adapted GoogLeNet to classify vibration response spectrograms and to identify the location and intensity of damage. The model uses continuous wavelet transforms to generate time-frequency spectrograms from train-borne acceleration signals at varying speeds, followed by image resizing to standardize input dimensions for the 2-D GoogLeNet architecture, enabling the classification of vibration responses across different speeds. In \cite{corballyDeeplearning2024}, an auto-calibrated vehicle-bridge interaction model is proposed to generate training data for damaged bridge scenarios and to train a CNN to classify damage type and location. The frequency spectrum from a laboratory-scale vehicle-bridge interaction model was interpolated and resampled at 0.05 Hz intervals using a cubic spline function to standardize input sizes. The algorithm accurately detected seized bearings and cracks in bridge beams, though crack location accuracy decreases at lower damage levels. In our previous work \cite{riahisamaniBidirectionalLongShortTerm2025}, we developed an LSTM-BiLSTM network to estimate track stiffness parameters from on-board vibration signals. The model was trained on simulated vehicle-track interactions at a single speed in the time domain and demonstrated accurate stiffness parameter estimation. It was also highlighted that the importance of incorporating temporal relationships in feature extraction and health condition estimation architectures of deep learning models for on-board vibration response analysis \cite{riahisamaniBidirectionalLongShortTerm2025}.

Most deep learning frameworks, particularly those based on convolutional neural networks (CNNs), require fixed input sizes. However, in on-board measurements, varying operational speeds result in signals of different lengths and spatial sampling frequencies for the same infrastructure segment. This necessitates preprocessing steps such as spectrogram generation, followed by spectrogram resizing or resampling, which can introduce limitations. Generating a spectrogram involves careful selection of hyperparameters, such as window type, length, and stride, all of which can influence the performance of the model. Additionally, the resizing or resampling of spectrograms may lead to information loss or the creation of synthetic data points, which can degrade the resolution of spectral features and affect overall accuracy. To address these challenges, the current paper proposes a fully automated methodology that processes data directly in the time domain. This approach effectively handles the variable-length inputs caused by different measurement speeds. Papers \cite{michauFullyLearnableDeep2022, frusqueRobust2024a} already introduced DeSpaWN and learnable wavelet packet transform (LWPT) networks for unsupervised high-frequency signal reconstruction, which can accommodate variable-length inputs. However, these methods do not specifically address the root cause of the variations, measurement speed. In contrast, our supervised learning framework not only incorporates wavelet-based analysis but also explicitly integrates measurement speed into the analysis.


Wavelet-inspired deep learning architectures are increasingly utilized to integrate spectral information into models across various fields. For example, \cite{liuMultilevel2019} introduced a multilevel Wavelet-CNN for image restoration, leveraging the Discrete Wavelet Transform for downsampling and the Inverse Wavelet Transform for up-sampling. In \cite{liWaveletKernelNet2022}, WaveletKernelNet CNN architectures were proposed for industrial diagnosis, where a continuous wavelet convolutional layer replaces the first standard CNN convolutional layer. This can provide a physically meaningful input layer, improving training performance and model accuracy \cite{liWaveletKernelNet2022}. Wavelet-informed deep learning frameworks have been applied to a variety of tasks in several application fields, such as ECG signal classification \cite{bounyEndtoEnd2020}, image classification \cite{yaoWaveViTUnifyingWavelet2022}, and time series analysis \cite{wangMultilevel2018}. These papers demonstrate that incorporating spectral analysis techniques, such as wavelet transforms, can enhance deep learning performance by providing physically meaningful information derived from wavelet decompositions. \rev{In the current paper, we propose a novel WaveletInception which integrates spectral information in the early stages via an LWPT module and ensures multi-scale feature extraction with 1D Inception-ResNet modules. We further employ BiGRU networks to account for operational conditions, particularly measurement speed, and to capture temporal dependencies in the analysis of vibration response signals. Sections \ref{preli} and \ref{Method} provide methodological preliminaries and further details.}

\section{Preliminaries}\label{preli}

\subsection{Discrete Wavelet Transform}

The wavelet transform is a powerful mathematical tool for analyzing signals in the time and frequency domains \cite{mallat1999wavelet}. Unlike the Fourier transform, which provides only a frequency-domain representation, the wavelet transform offers a multi-resolution analysis by decomposing signals into components at various scales. This capability makes it particularly effective for analyzing non-stationary signals, such as on-board vibration signals.

The Discrete Wavelet Transform (DWT) is a computationally efficient version of the wavelet transform, widely used for signal and image processing. The basic operation of DWT is the convolution operation of a low-pass and a high-pass filter, followed by a downsampling by a factor of two, which can be denoted as

\begin{equation}\label{DWT}
\begin{cases}
    (x^\mathrm{l})_n = (f^\mathrm{l} *  x )_{2n}\\
    (x^\mathrm{h})_n = (f^\mathrm{h} *  x )_{2n}
\end{cases}
\end{equation}

\noindent where $x^\mathrm{l}$ and $x^\mathrm{h}$ correspond to the low-pass (approximation) and high-pass (details) filtered input data with a downsampling rate of 2, $f^\mathrm{l}$ and $f^\mathrm{h}$ are low-pass and high-pass filters, $*$ is the convolution operation, $x$ is the input signal with the signal length of $2n$. The perfect reconstruction and orthogonality property of the DWT ensure that all information is retained in its coefficients despite the downsampling \cite{liuMultilevel2019, mallat1999wavelet}.

\subsection{Discrete Wavelet Packet Transform and LWPT}

\begin{figure}[thpb]
      \centering
       \centering\includegraphics[scale=0.8,trim=0 40 110 0,clip]{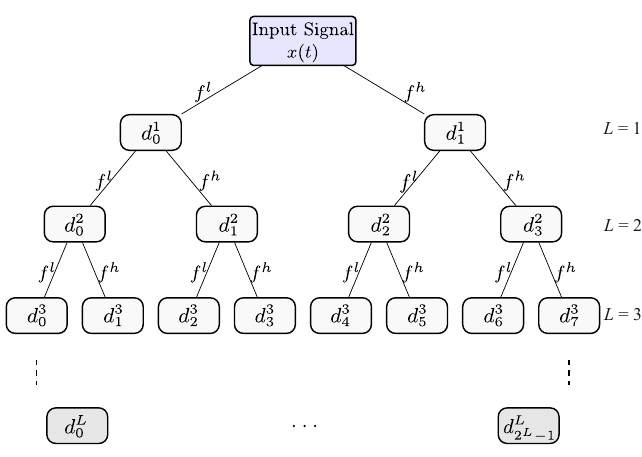}
      \caption{Tree-like structure of the Wavelet Packet Transform (WPT).}
      \label{WPT}
   \end{figure}

The Discrete Wavelet Packet Transform (WPT) provides a multi-resolution analysis by decomposing a signal into uniform frequency bands through a hierarchical, tree-like structure \cite{mallat1999wavelet}. At each level $L$, the signal is convolved with low-pass and high-pass filters (Eq.~\ref{DWT}), splitting the spectrum into finer sub-bands. While this process doubles the frequency resolution at each step, it reduces the temporal resolution by half. The result is a comprehensive decomposition where each leaf node in the WPT tree (see Figure \ref{WPT}) represents a specific frequency interval.

\rev{The Learnable Wavelet Packet Transform (LWPT), originally introduced in \cite{michauFullyLearnableDeep2022, frusqueRobust2024a}, initializes its filters using standard DWT bases, such as Haar or Daubechies ($db4$), but refines them dynamically via backpropagation1. Unlike traditional WPT, which relies on fixed coefficients that may not align with the complex transients of railway vibrations, LWPT learns the optimal filter bank directly from the data. This approach leverages the perfect reconstruction and orthogonality of wavelets to downsample high-frequency vibration signals without the information loss typical of standard pooling layers \cite{michauFullyLearnableDeep2022, frusqueRobust2024a}.  Furthermore, the LWPT incorporates an adaptive denoising mechanism that exploits coefficient sparsity; by utilizing a differentiable double sharp sigmoid activation function, the model applies learnable thresholds to suppress stochastic noise while preserving salient structural transients \cite{michauFullyLearnableDeep2022}. Compared to the DWT and WPT, this adaptive capability ensures that refined, noise-reduced mechanical signatures are effectively propagated through the network.}

\rev{In the current paper, the LWPT module is coupled with a $1 \times 1$ channel-projection layer to adapt the spectral output to a supervised learning framework effectively. Furthermore, the model is evaluated both with and without the denoising function to empirically assess the impact of adaptive thresholding on health-condition estimation.}

\subsection{1DCNN}

A Convolutional Neural Network (CNN) is a deep learning architecture designed to extract spatial and temporal features from input data using convolutional operations with filter banks \cite{aggarwal2018neural}. While CNNs are commonly associated with image data, their 1D variants are particularly effective for processing sequential or time-series data, such as vibration and audio signals. A 1D CNN operates on one-dimensional signals, with convolutional kernels in a filter bank sliding along the temporal dimension. These networks are designed to capture local patterns or dependencies in the input signal. This process generates feature maps that capture abstract representations of the input and highlight different aspects of the input data. The formula for a typical convolutional layer is



\begin{equation}
    y_f = \mathrm{conv1D}(W_{f},x) + b_f
    \label{eq:hidout}
\end{equation}

\noindent
where $y_f$, $W_f$, and $b_f$ denote the output vector, weight vector, and bias parameter of filter $f$, respectively; $x$ is the input time sequence; and $\mathrm{conv1D}$ represents the one-dimensional convolution operator. The $i^{\text{th}}$ output of the convolution is computed as:

\begin{equation}
    [\mathrm{conv1D}(W_f, x)]_i = [W_f * x]_i = \sum_{j=0}^{N_f - 1} w_{f,j} \cdot x_{i + j}
    \label{eq:hidoutt}
\end{equation}

\noindent where $N_f$ is the length of the filter $f$, $w_{fj}$ is the $j^{th}$ element of the vector $W_f$, and $i$ is the starting index in the input for each convolution window.

\subsection{\rev{Gated Recurrent Unit (GRU)}}

\rev{Gated Recurrent Units (GRUs) are a streamlined variant of Recurrent Neural Networks designed to capture temporal dependencies while mitigating the vanishing gradient problem \cite{cho2014learning, chungEmpiricalEvaluation2014a}. Unlike the LSTM, the GRU achieves an efficient information flow using a simplified architecture that merges the forget and input gates into a single update gate and utilizes a reset gate to regulate the integration of previous hidden states.}

\rev{Given an input sequence $\{ x_t \}_{t=1}^T$, the GRU updates its hidden state $h_t$ at each time step $t$ by combining the current input $x_t$ with the previous hidden state $h_{t-1}$. The update gate determines how much of the previous information is carried over to the current hidden state, while the reset gate decides how much of the past context should be discarded when computing the new candidate state. This mechanism allows the model to retain long-term dependencies effectively with fewer parameters than traditional LSTMs.}

\rev{While a standard GRU processes data in a forward direction, a Bidirectional GRU (BiGRU) adds a backward pass, capturing both past and future signal context. This comprehensive temporal perspective is especially valuable for vibration-based infrastructure health monitoring \cite{riahisamaniBidirectionalLongShortTerm2025}.}

\subsection{Multi-modal data fusion}

Multi-modal data fusion combines data from different modalities, such as vibration signals, LiDAR, audio, and speed measurements, each of which provides complementary insights into the same phenomenon. Integrating multiple data sources can result in more robust, accurate, and generalizable models \cite{lahat2015multimodal}. Existing multi-modal fusion methods can be categorized into three main types: feature-level fusion, decision-level fusion, and hybrid fusion \cite{lahat2015multimodal}. Feature-level fusion entails extracting features from each modality and merging them into a unified representation. In contrast, decision-level fusion combines the outputs of separate models trained on individual modalities. Hybrid fusion integrates feature and decision levels, leveraging the strengths of each approach \cite{lahat2015multimodal}. 

Feature fusion is widely used across various domains to exploit correlations among multiple input modalities, sensors, or extracted features. This method leverages inherent correlations among features across modalities, thereby enhancing training. However, it does face challenges related to temporal synchronization, as features from closely related modalities may be sampled at different time instants \cite{ZHANG2019158}. For instance, in \cite{dangDataDriven2021}, a hybrid 1D CNN-LSTM network is proposed that combines autoregressive features, discrete wavelet features, and empirical mode decomposition features from multiple sensors into a unified representation for model training. Similarly, \cite{mouDriver2021} utilized an attention mechanism to integrate features extracted by a CNN-LSTM network from three distinct input sources: environmental data, vehicle data, and driver eye data. 

In the current paper, we propose a feature-fusion approach that combines features from vibration-response signals and operational-condition embeddings. We use a BiGRU layer to capture temporal and interrelated dependencies within the feature-extraction network, as explained below.

\section{WaveletInception-BiGRU Networks}\label{Method}

\subsection{Problem statement}

\rev{Effective Structural Health Monitoring (SHM) via on-board vibration signals requires multi-resolution analysis to identify defects across various frequency bands \cite{shenEvaluatingRailwayTrack2023a,lamprea-pinedaRailwayTrackReceptance2024}. Standard CNNs, while excellent at spatial pattern recognition, lack an inherent mechanism for multi-resolution spectral decomposition, and their effectiveness in performing spectral analysis is still not well understood \cite{bounyEndtoEnd2020}. Furthermore, real-world measurement conditions—specifically varying measurement speeds—induce non-stationarity in vibration signals, leading to inconsistent excitation levels and varying signal lengths. Conventional preprocessing, such as fixed-ratio resampling or rigid resizing, often leads to information loss or fails to account for the physical relationship between measurement speed and vibration intensity. Additionally, most existing infrastructure health monitoring frameworks provide aggregated health indices over long segments (e.g., 30–200 m) \cite{unsiwilaiMultipleaxle2023, unsiwilaiEnhanced2024}. Such coarse resolution hinders localized maintenance and component-level defect identification.}

\rev{To address these challenges, we propose the WaveletInception-BiGRU framework, which integrates three key components: a WaveletInception module for learnable spectral decomposition and multi-scale feature extraction from vibration signals; a feature fusion technique to incorporate operational conditions; and a BiGRU network for sequential modeling and capturing temporal relationships. This combination enables high-resolution, component-level health estimation by leveraging both physical and operational insights.}

\subsection{WaveletInception-BiGRU}

\rev{The proposed architecture consists of two main stages: (I) a WaveletInception stage for multi-scale spatial feature extraction, and (II) a BiGRU-based temporal head for multi-modal fusion and sequential health conditions estimation. This design handles variable-length inputs while maintaining a physically meaningful relationship between the signal features and the infrastructure's spatial layout. Figure \ref{fg:method} shows an overview of the proposed WaveletInception-BiGRU networks}

\subsubsection{WaveletInception Feature Extraction Networks}

\rev{The WaveletInception stage serves as the primary feature extractor, designed to handle the high sampling rates characteristic of vibration signals. It comprises an LWPT followed by a channel projection layer, and Inception-ResNet networks.}

\rev{The WaveletInception network begins with an LWPT stem that hierarchically decomposes the raw vibration signal into $2^L$ sub-bands (where $L$ denotes the decomposition level). This process utilizes learnable high- and low-pass filters to extract multi-resolution features while simultaneously downsampling the temporal dimension by a factor of two at each level. Thanks to its orthogonality, LWPT enables high-frequency signals to be downsampled without information loss \cite{michauFullyLearnableDeep2022}. The filters are initialized with Discrete Wavelet Transform (DWT) bases, such as Haar or Daubechies db4 \cite{frusqueRobust2024a}, and are then refined through backpropagation. This initialization provides a physically grounded starting point for time-frequency decomposition, where db4 filters, in particular, offer smoother signal boundary handling for zero-padded regions compared to Haar filters.}

\rev{Since the diagnostic relevance of wavelet sub-bands varies significantly across different operational conditions, a $1\times 1$ convolution layer is implemented as a learned channel projection. This layer replaces manual sub-band selection by serving as an automated weighting mechanism that identifies and prioritizes the most informative spectral components while suppressing noisy channels. By learning these projections during training, the network adaptively optimizes the integration of disparate frequency information. Following this projection, two sequential 1D convolutional layers further refine the spatial-spectral features and downsample the temporal dimension. This process optimizes the feature density and ensures that only the most salient local patterns are propagated to the subsequent 1D Inception-ResNet backbone.}

\begin{figure}[thpb]
      \centering
      \includegraphics[scale=0.73, trim=0 20 0 220,clip]{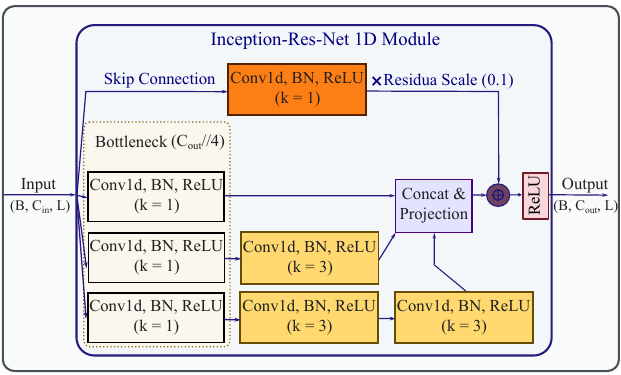}  
      \caption{1D Inception block structure.}
      \label{inceptionfig}
   \end{figure}

\rev{The 1D Inception-ResNet network efficiently extracts high-level features from on-board vibration signals. As shown in Figure~\ref{inceptionfig}, these blocks utilize parallel convolutional branches to capture varied signal patterns while leveraging residual connections to stabilize training in deep architectures. To minimize computational overhead, bottleneck layers are integrated, which effectively refine fine-grained features without excessive parameterization. Unlike conventional 2D models that require complex signal-to-image transformations, these 1D modules operate directly on the temporal axis. This not only improves efficiency but also ensures physical interpretability, as the learned filters act as adaptive digital signal processors that correspond to the time-domain characteristics \cite{borghesaniFourierbasedExplanation2023} of the on-board vibration response.}

\rev{The hyperparameters of the WaveletInception networks consist of the number of decomposition levels ($L$) for the LWPT, the number of Inception-ResNet blocks, and the number of output channels of the convolution layers. Table~\ref{tab:tensor_shapes} shows the fine-tuned WaveletInception networks after hyperparameter optimization for the two case studies conducted.}

\subsubsection{BiGRU Networks, Feature Fusion and Temporal Modeling}

\rev{Our proposed BiGRU networks consist of two BiGRU modules as shown in Figure~\ref{fg:method}. First, the BiGRU module extracts bidirectional temporal features from the concatenated vibration signal features and the operational-condition embedding, in particular, the speed embedding. The vibration features are the outputs of the WaveltInception network, and the measurement speed is processed through a lightweight Multi-Layer Perceptron (MLP). This transforms the scalar speed into a high-dimensional speed embedding that matches the temporal length of the vibration features.}

\rev{The first BiGRU layer fuses the vibration features with the speed embedding. This layer integrates temporal dependencies into the feature-extraction process and captures interrelationships in the concatenated features of the vibration signal and operational conditions, such as signal length variation and excitation level. Additionally, the BiGRU layer accounts for the relationships among measurement speed, signal length, and padding length, thereby further refining the learned features.}

\rev{To achieve component-level resolution, we implement a distributed sampling strategy designed to extract $N$ feature vectors from the BiGRU outputs for sequential modeling. In this approach, both the value of $N$ and the specific indices for feature extraction are determined by the physical layout of the infrastructure. For instance, if an infrastructure segment contains 30 equidistant sleepers, $N$ is set to 30, and feature vectors are extracted at uniform intervals from the temporal sequence to align with the sleeper positions. This physical-temporal mapping ensures that each network output corresponds to a distinct structural component. By incorporating prior knowledge of the infrastructure configuration, the model maintains a one-to-one correspondence with the physical layout, thereby significantly improving the accuracy of high-resolution estimation. The selected feature vectors pass through a dropout layer before being fed into the final temporal modeling in the health estimation head.}

Finally, these selected vectors pass through a dropout layer prior to temporal modeling in the health estimation head.



\rev{The second BiGRU module serves as a sequential model for the estimation head in monitoring infrastructure health conditions. In this context, sequential modeling captures spatial dependencies across a sequence of beams along the infrastructure, with each beam corresponding to a single time step. The BiGRU layer processes the input at each time step and captures temporal dependencies along the sequences. The BiGRU outputs are passed through fully connected layers to map the learned representations onto specific health-condition indicators, such as regression (as in Case Study I) and classification (as in Case Study II) tasks. This sequential modeling enables detailed predictions at the component level, corresponding to the structure's physical layout. Consequently, this method improves the resolution of health condition estimations and facilitates comprehensive assessment of individual components, providing valuable insights for maintenance planning.}

\rev{The proposed framework utilizes BiGRU to capture comprehensive temporal context by concatenating hidden states from both forward and backward passes. Physically, the health condition of infrastructure elements—like railway sleepers or bridge beams—is characterized not only by their local dynamic response but also by mechanical signatures propagated through adjacent components. This mechanism enables the model to identify complex, long-range dependencies between fused vibration features and operational conditions, such as the measurement speed. By integrating multi-modal inputs from both preceding and succeeding spatial steps, the BiGRU layers effectively capture the global context inherent in on-board vibration data. Consequently, this bidirectional awareness markedly improves estimation performance relative to unidirectional models, which fail to incorporate future signatures from downstream vibration signals to refine the current health-condition assessment.}

\rev{The hyperparameters of the temporal modeling and estimation head network include the size of the operational condition embedding module, the number of BiGRU units in the BiGRU layer, the size of the fully connected layers, and the dropout ratio. Table~\ref{tab:tensor_shapes} shows the parameters fine-tuned in the two case studies.}

\begin{figure}[H]
      \centering
       \includegraphics[scale=0.65, trim=0 150 0 290,clip]{Figures/drawing_methodology3.pdf}  
      \caption{Overview of the WaveletInception-BiGRU architecture, including WaveletInception vibration signal feature extraction, operational conditions feature fusion, and BiGRU sequential health conditions estimation.}
      \label{fg:method}
   \end{figure}

\rev{\section{Illustrative case study of railway track health monitoring}\label{case}}

\rev{The Axle-Box Acceleration (ABA) measurement system is a cost-effective drive-by inspection method used on operational trains to capture vibration signals through accelerometer sensors mounted on axle boxes \cite{phusakulkajornHybridNeuralModel2025, molodovaAutomaticDetectionSquats2014}. This system allows for the development of a fully automated framework to analyze ABA vibrations, enhancing railway infrastructure monitoring over time and space.}

\rev{To demonstrate this methodology, we present two case studies: (I) railway track stiffness estimation and (II) railway transition zone identification. In both cases, time-domain ABA signals serve as inputs to our deep learning architecture, which integrates preprocessing, feature extraction, and estimation.}

\rev{\subsection{Case Study I: Railway track stiffness estimation}}

\rev{Railway track stiffness is a key indicator of infrastructure health, primarily influenced by the stiffness of the ballast substructure and fastening components like bolts, clamps, and rail pads. Estimating the stiffness of these components offers insights into track condition. For instance, defects in railway sleepers, such as hanging sleepers or ballast crushing, can significantly reduce track stiffness \cite{shiCritical2023}.} 

\rev{In this paper, we use simulated ABA vibration responses to estimate track stiffness parameters, characterized by rail pad stiffness and ballast stiffness, represented as the vector \( k = [k_{\mathrm{p}}, k_{\mathrm{b}}]^\intercal \).}

\rev{\subsubsection{Data}}

\begin{figure}[H]
      \centering
      \framebox{\parbox{\linewidth}{%
      \includegraphics[scale=0.73,trim=6 10 0 0,clip=true ]{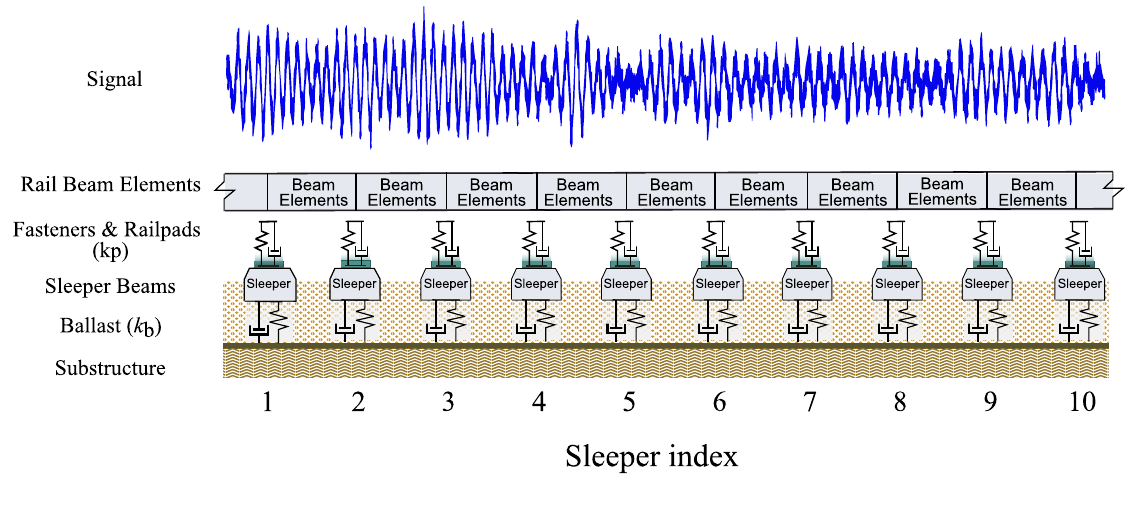}}}
         
      \caption{The layout of the 10-sleeper track segment.}
      \label{fg:layout}
\end{figure}

\rev{We simulate vehicle-track interactions over segments of 10 sleepers at speeds of 35 km/h, 50 km/h, 65 km/h, 80 km/h, and 90 km/h, under four stiffness variation scenarios. The dataset is generated using a finite-element vehicle-track interaction model implemented in MATLAB, where the rail and sleepers are modeled as Timoshenko beam elements supported by rail pads and ballast with stiffness and damping properties. The vehicle is represented as a wheelset with Hertzian contact for wheel-rail interaction. Further details on the model can be found in \cite{shenEvaluatingRailwayTrack2023a}. Figure \ref{fg:layout} shows a schematic of the track segment and the corresponding ABA measurements for each combination of speed and stiffness scenario.
}

The four stiffness scenarios include:

(I) Uniform track stiffness: Normal conditions with stiffness values randomly sampled from either the nominal (R1) or moderate (R2) stiffness ranges in Table \ref{table_range}.

(II) Local stiffness reduction in one sleeper: Simulates a defect like a hanging sleeper, with stiffness values set to R3 representing varying fault severities.

(III) Local stiffness reduction in three sleepers: Models reduced support over a longer section, with values sampled uniformly from R1 to R2 or R3.

(IV) Transition zones: Simulates stiffness variations between the two halves of segments, with values randomly sampled from R1 and R2.

\begin{figure}[H]
      \centering
      \framebox{\parbox[c][14cm]{\linewidth}{%
      \includegraphics[scale=0.7,trim=0 290 40 50,clip=true]{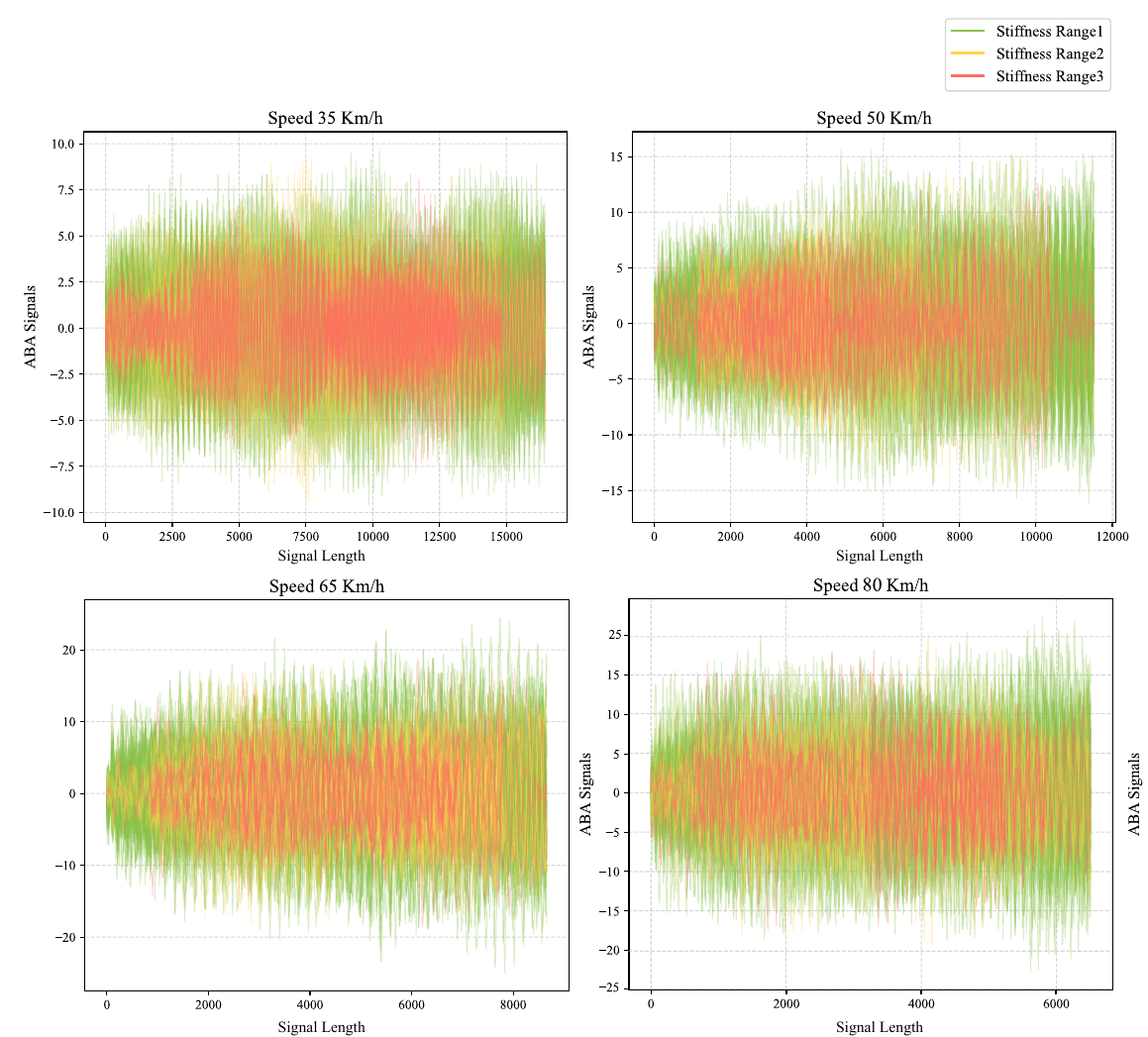}}}
         
      \caption{Representation of the simulated ABA signals in four different measurement speeds. ABA signals obtained over the stiffness range sets R1, R2, and R3 are plotted in green, yellow, and red, respectively.}
      \label{fg:signals_tm}
\end{figure}

\begin{figure}[H]
      \centering
      \framebox{\parbox[c][14cm]{\linewidth}{%
      \includegraphics[scale=0.7,trim=20 280 0 20,clip=true]{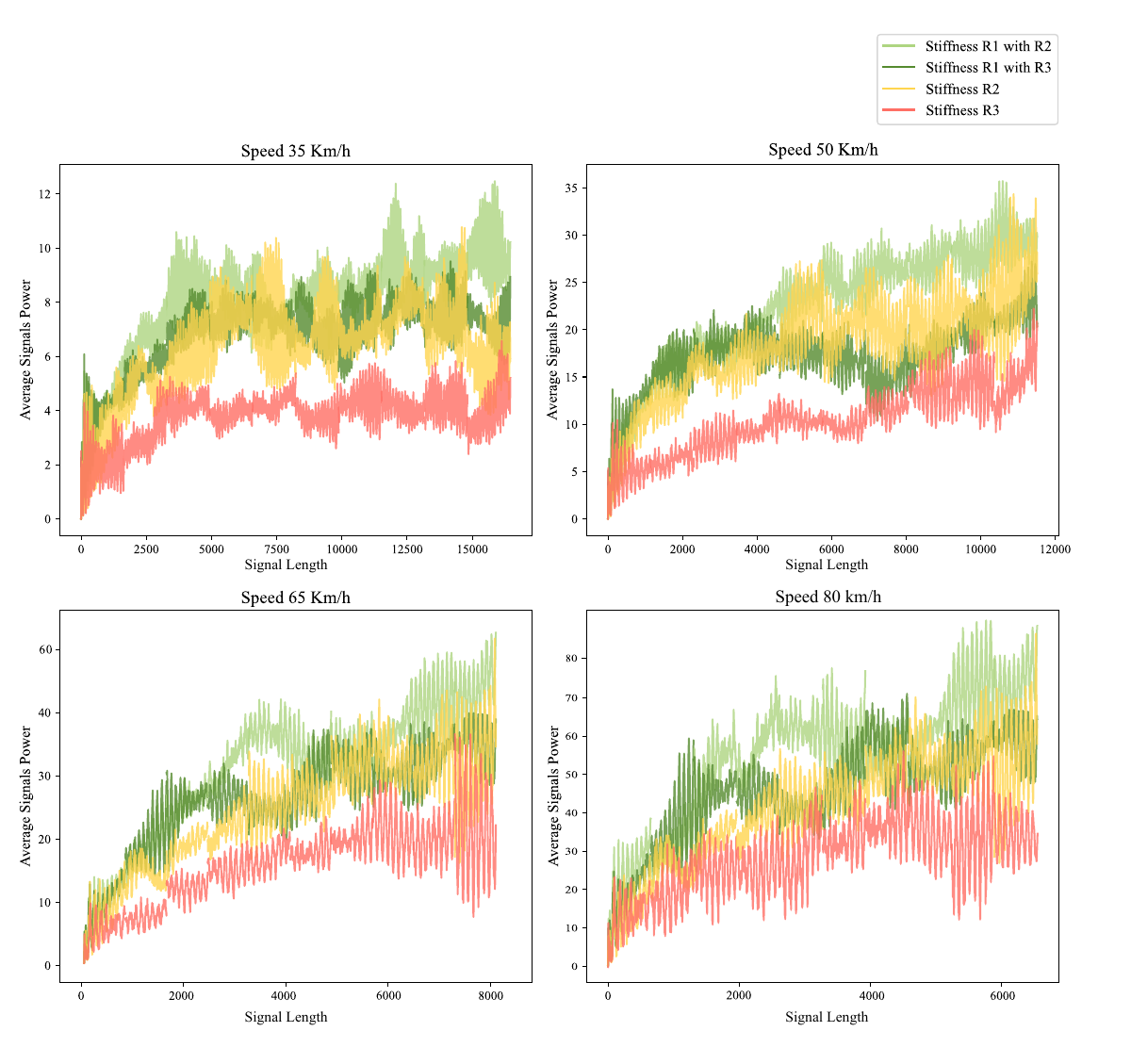}}}
         
      \caption{ Impact of local stiffness reduction on long-range ABA signal power, showing average signal power for three sleepers with stiffness range R1 reduced to ranges R2 (light green) and R3 (dark green), alongside original ranges R2 (yellow) and R3 (red).}
      \label{fg:ave_power}
\end{figure}

\rev{Figure \ref{fg:signals_tm} illustrates the ABA signals across stiffness ranges (R1, R2, R3) and speeds. Lower stiffness correlates with lower signal amplitude, whereas higher speeds increase signal amplitude. Figure \ref{fg:ave_power} illustrates the average signal power across different stiffness ranges, specifically in the context of localized stiffness reductions at three sleepers. The figure focuses on stiffness range R1 while examining adjacent reductions in either range R2 (light green) or range R3 (dark green). The figure shows that a local reduction in stiffness affects the ABA signals recorded in nearby sections, even when those sections remain within the stiffness range R1. This observation supports our choice of methodology: employing a sequential BiGRU-based model to capture long-range dependencies within the data effectively.}

\rev{The dataset comprises 12\,500 records, with 2\,500 for each measurement speed across different scenarios. To simulate real-world conditions, Gaussian noise, modeled as \(\sim N(0, \sigma^2)\), is added to the ABA signals, achieving a noise-to-signal ratio of 5\%. This noise variance accounts for 5\% of the ABA signal power, replicating measurement noise from environmental factors and vehicle disturbances \cite{gonzalezEffective2023, malekjafarianMachine2019}. The dataset is divided into training (70\%), validation (15\%), and testing (15\%) sets \cite{bishopDeepLearning2023}.}

\begin{table}[H]
\centering
\renewcommand{\arraystretch}{1.25}
\caption{Range of parameter values used for railpad and ballast stiffness.}
\label{table_range}
\fontsize{8.5}{10.5}\selectfont
\begin{tabularx}{\textwidth}{>{\raggedright\arraybackslash}p{12mm} 
                                   >{\centering\arraybackslash}X 
                                   >{\centering\arraybackslash}X}
\toprule
Range sets & $k_\mathrm{p}\, (\mathrm{N/m})$ & $k_\mathrm{b}\, (\mathrm{N/m})$ \\
\midrule
R1 & $2.0\cdot 10^8$\,--\,$3.0\cdot 10^8$ & $1.6\cdot 10^7$\,--\,$2.2\cdot 10^7$ \\
R2 & $1.0\cdot 10^8$\,--\,$2.0\cdot 10^8$ & $1.0\cdot 10^7$\,--\,$1.6\cdot 10^7$ \\
R3 & $0.1\cdot 10^8$\,--\,$1.0\cdot 10^8$ & $0.4\cdot 10^7$\,--\,$1.0\cdot 10^7$ \\
\bottomrule
\end{tabularx}
\end{table}

\rev{\subsection{Case Study II: Transition zones identification using real-world data}}

\rev{Transition zones near viaducts, bridges, tunnels, and level crossings are critical parts of railway infrastructure. They often experience abrupt changes in track stiffness, e.g., when shifting from ballast to slab track \cite{phusakulkajornHybridNeuralModel2025a, wangExperimentalAnalysisRailway2018, wangMethodologyComprehensiveAnalysis2018}. These stiffness variations can lead to increased dynamic loads, differential settlements, and accelerated track degradation, posing challenges for maintenance and safety \cite{wangExperimentalAnalysisRailway2018}. Accurate identification and monitoring of these zones are vital for the reliability of railway networks. In this case study, we apply our proposed framework to pinpoint regions within transition zones at a sleeper-level resolution, using real-world measurements of ABA signals collected at different speeds.}

\rev{\subsubsection{Data}}

\begin{figure}[H]
      \centering
      \framebox{\parbox[c][7.2cm]{10cm}{%
      \includegraphics[scale=1.4,trim=285 0 0 80,clip=true]{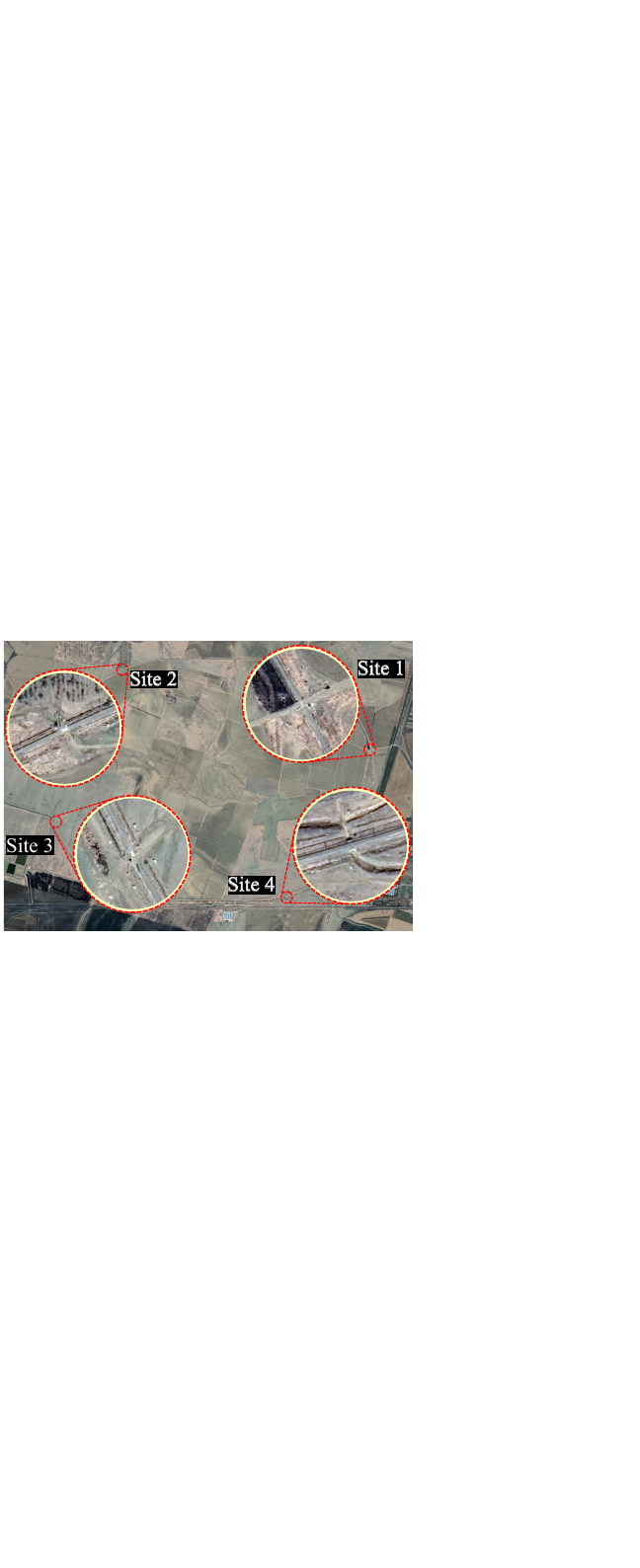}}}
         
      \caption{Case Study II map, four transition zone sites selected from the Romania test ring. Maps are adapted from Google Maps 2025.}
      \label{case2}
\end{figure}

\rev{This case study uses data from the Făurei testing ring in Romania, operated by the Romanian Railway Authority (AFER). ABA measurements are collected over transition zones near crossing levels and viaducts during dynamic testing at speeds ranging from 20 km/h to 140 km/h. Figure \ref{case2} shows the test ring map and the four selected transition zones for this paper.}

\rev{The ABA system on the measurement train collects 16 vibration responses from the two wheelsets during each round. These signals are extracted over 18-m track segments, focusing specifically on transition zones. Data from sites 1–3 are utilized for model training (85\%) and validation (15\%), totaling 7,000 ABA samples. To rigorously assess generalization, site four is reserved as an entirely unseen test set. This set, comprising 107 transition-zone and 110 normal-track samples, is used exclusively for final evaluation and deployment. Such a separation ensures an unbiased estimate of model performance \cite{bishopDeepLearning2023}.
}

\begin{figure}[H]
      \centering
      \framebox{\parbox[c][10.5cm]{\linewidth}{%
      \includegraphics[scale=0.4,trim=0 0 0 0,clip=true]{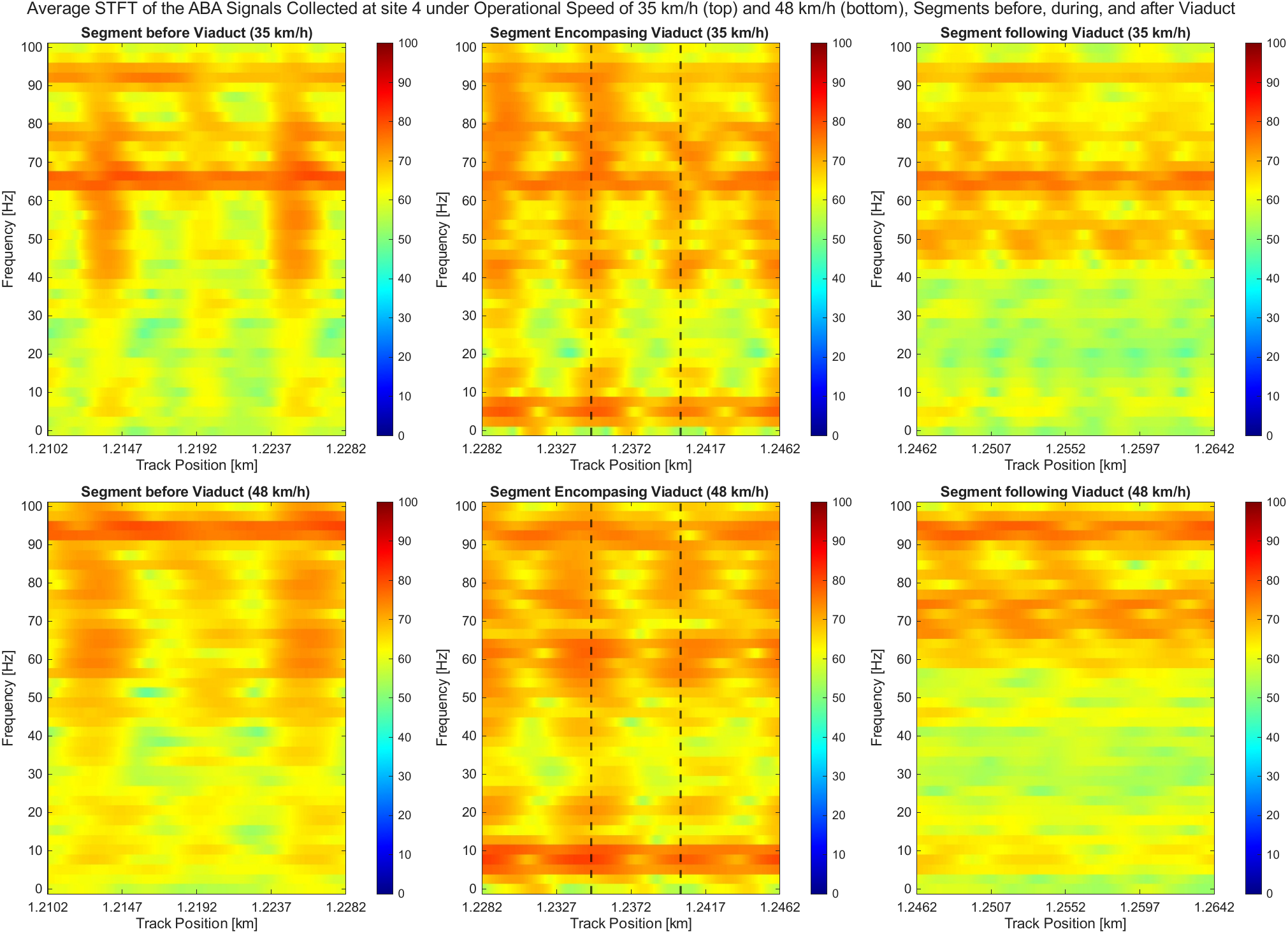}}}
         
      \caption{Influence of the local change in track stiffness on a longer range of SFTF spectrogram in the low frequency. The viaduct is located between 1.2348 km and 1.2402 km.}
      \label{case2SFTF}
\end{figure}

\rev{Figure \ref{case2SFTF} presents the average short-time Fourier transform (STFT) spectrograms of the ABA signals from three consecutive segments, before, during, and after the viaduct, at Site 4. The left spectrogram covers the interval from 1.2102 km to 1.2282 km. The middle segment includes the viaduct located between 1.2348 km and 1.2402 km. The right segment extends from 1.2462 km to 1.2642 km.}

\rev{Each spectrogram illustrates the changes in vibration frequency content below 100 Hz. The plots indicate that the dominant frequency components associated with variations in substructure stiffness are primarily concentrated around 10 Hz. This frequency range in the figures representing conditions before and after the viaduct suggests that changes in track stiffness can influence ABA responses over a distance, highlighting the presence of long-range spatial dependencies.}

\rev{These findings highlight the necessity of modeling ABA signals as sequential data with long-range dependencies, which allows the learning framework to capture both localized features and their broader impact on the characteristics of ABA signals along the track.}

\rev{\subsection{Training, validation, and test design}}

\begin{figure}[thpb]\label{fg:learning}
      \centering
      \framebox{\parbox{2.8in}{%
      \includegraphics[trim=0cm 3.5cm 0cm 0cm, clip, scale=0.68]{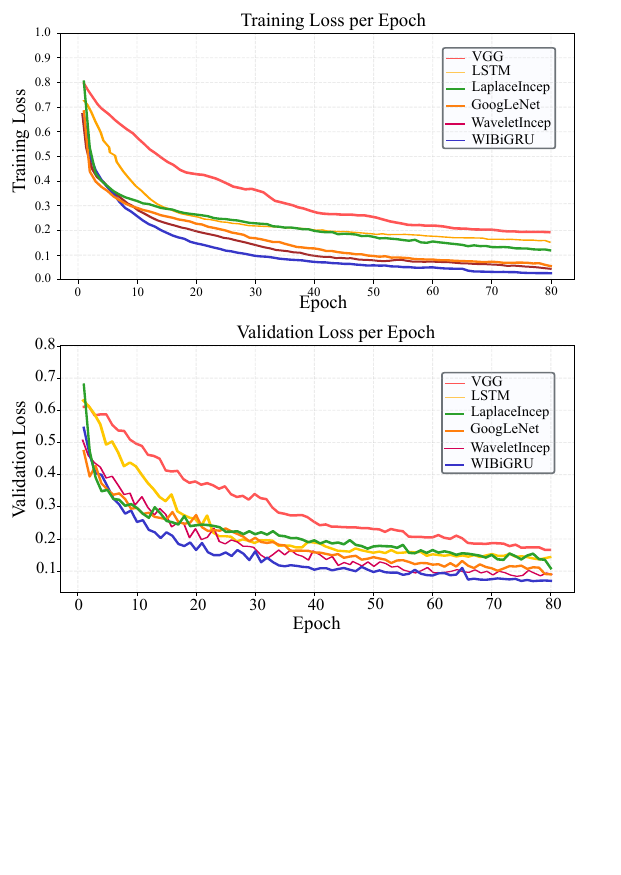}
      }}   
      \caption{Training and validation Loss convergence for the proposed model and baselines.}
      \label{fgepochs}
   \end{figure}

\rev{Fixed training, validation, and test sets are used to evaluate the models. The mean squared error (MSE) loss for Case Study I and the binary cross-entropy with logits loss (BCEWithLogitsLoss) for Case Study II are minimized using the Adam optimizer \cite{kingmaAdam2017}. To enhance loss convergence, we implemented a learning rate warm-up, followed by a scheduler with a factor of 0.5 and a patience of 6. For example, Figure~\ref{fgepochs} shows the loss convergence of the models on the training and validation sets in Case Study I. We tuned hyperparameters—including the number of layers, units, dropout rates, and learning rates—during the validation process. In particular, random search in Weights\& Biases \cite{wandb} is used to explore 50 hyperparameter configurations for each architecture, efficiently covering high-dimensional spaces without the combinatorial cost of exhaustive grid search \cite{bergstraRandomSearch2012}. Table~\ref{tab:tensor_shapes} shows the configuration of the fine-tuned architectures for both case studies.}

\begin{table}[H]
\caption{Wavelet–Inception–BiGRU architecture fine-tuned for Case Studies I and II.}
\label{tab:tensor_shapes}
\renewcommand{\arraystretch}{1.25}
\fontsize{8.5}{10.5}\selectfont
\centering

\begin{tabularx}{\linewidth}{
    >{\raggedright\arraybackslash}p{4.4cm}
    >{\centering\arraybackslash}X
    >{\centering\arraybackslash}X
}
\toprule
\multirow{2}{*}{\textbf{Network stage}} &
\multicolumn{2}{c}{\textbf{Output tensor size (Batch $\times$ Channels $\times$ Length)}} \\
\cmidrule(lr){2-3}
& \textbf{Case Study I} & \textbf{Case Study II} \\
\midrule

Input vibration signal
& $32 \times 1 \times 16460$
& $32 \times 1 \times 78150$ \\

LWPT stem (level $L$)
& $32 \times 64 \times 258$
& $32 \times 128 \times 611$ \\

Channel projection (Conv1D, $k=1$)
& $32 \times 96 \times 258$
& $32 \times 128 \times 611$ \\

Conv1D$_1$
& $32 \times 96 \times 258$
& $32 \times 96 \times 306$ \\

Conv1D$_2$
& $32 \times 96 \times 258$
& $32 \times 96 \times 306$ \\

Inception–ResNet module 1
& $32 \times 64 \times 258$
& $32 \times 128 \times 306$ \\

Inception–ResNet module 2
& $32 \times 98 \times 258$
& $32 \times 128 \times 306$ \\

Max pooling
& $32 \times 98 \times 128$
& $32 \times 128 \times 152$ \\

Inception–ResNet module 3
& $32 \times 160 \times 128$
& $32 \times 256 \times 152$ \\

Inception–ResNet module 4
& $32 \times 256 \times 128$
& $32 \times 380 \times 152$ \\

Max pooling
& $32 \times 256 \times 63$
& $32 \times 380 \times 75$ \\

Speed fusion (channel concatenation)
& $32 \times 320 \times 63$
& $32 \times 444 \times 75$ \\

Temporal modeling (BiGRU$_1$)
& $32 \times 256 \times 63$
& $32 \times 256 \times 75$ \\

Uniform temporal sampling
& $32 \times 256 \times 10$
& $32 \times 256 \times 30$ \\

Temporal modeling (BiGRU$_2$)
& $32 \times 196 \times 10$
& $32 \times 192 \times 30$ \\

Fully connected layer 1
& $32 \times 98 \times 10$
& $32 \times 96 \times 30$ \\

Fully connected layer 2 (output)
& $32 \times 2 \times 10$
& $32 \times 1 \times 30$ \\

\bottomrule
\end{tabularx}
\end{table}

\rev{A systematic comparison of the proposed WaveletInception-BiGRU model is conducted across external baselines and its own ablations. The analysis mainly consists of three key components: preprocessing (stem), speed-data fusion, and temporal feature extraction. In the preprocessing phase, several techniques are evaluated to handle variable-length input signals. These techniques include interpolation-based representations using the Continuous Wavelet Transform (CWT) and Short-Time Fourier Transform (STFT), as well as zero-padding the signals followed by the LWPT, STFT, and length-aware modules. For speed integration, two mid-level data fusion techniques, early and late fusion, are examined. Temporal feature extraction is compared across LSTM, GRU, BiLSTM, and BiGRU architectures. Our architecture—including a variant without auxiliary speed fusion—is benchmarked against literature baselines: VGG \cite{lockeUsing2020}, LSTM-BiLSTM \cite{riahisamaniBidirectionalLongShortTerm2025}, Laplace-Inception \cite{liWaveletKernelNet2022}, GoogLeNet \cite{hajializadehDeep2023}, and WaveletInception networks. Variants of LWPT, including Haar-WaveletInception and db4-WaveletInception, with and without the denoising module, are explored. Optimal hyperparameters are selected via validation, and test set performance is reported. For reproducibility, code is available on GitHub \cite{rezars9701_waveletinception_bigru_2025}. The Case Study I dataset is available online \cite{Vehicledata}; the Case Study II dataset is proprietary but available upon request from the corresponding author.}

\rev{\subsection{Results}}

\rev{Our analysis of the proposed model architecture begins with the stem module, which handles data preprocessing. This step is essential for analyzing drive-by vibration signals measured at different speeds, as varying signal lengths pose challenges for deep learning methods—particularly for CNNs, which require fixed-size inputs. To address this issue, Table~\ref{tab:abl_preprocessing} compares four preprocessing strategies: resizing the Continuous Wavelet Transform (CWT) and the Short-Time Fourier Transform (STFT) spectrograms to a standard size, zero-padding followed by the STFT, zero-padding followed by the LWPT stem module, and zero-padding combined with masking and length-aware components. The proposed zero-padding-based LWPT preprocessing achieves the best performance, yielding the lowest overall mean absolute percentage error (MAPE) of 5.56\% in Case Study~I and the highest classification accuracy of 93.29\% in Case Study~II. The length-aware LWPT variant ranks second, with an overall MAPE of 6.41\% in Case Study~I and an accuracy of 91.78\% in Case Study~II. In contrast, zero-padding followed by STFT exhibits the weakest performance, with the highest overall MAPE of 6.87 in Case Study~I and the lowest accuracy of 84.78\% in Case Study~II.}

\begin{table}[H]
\setlength{\tabcolsep}{2.5pt}
\renewcommand{\arraystretch}{1.25}
\fontsize{8.5}{10.5}\selectfont
\centering
\caption{Comparison of data preprocessing strategies for handling variable-length vibration signals in the proposed framework.}
\label{tab:abl_preprocessing}

\begin{tabularx}{\linewidth}{
    p{2.4cm}
    *{7}{>{\centering\arraybackslash}X}
}
\toprule
\multirow{2}{*}{\textbf{Model Types}} &
\multicolumn{3}{c}{\textbf{Case Study 1 (MAPE \%)}} &
\multicolumn{4}{c}{\textbf{Case Study 2 (Classification \%)}} \\
\cmidrule(lr){2-4} \cmidrule(lr){5-8}
& \textbf{$k_\mathrm{p}$}
& \textbf{$k_\mathrm{b}$}
& \textbf{Overall}
& \textbf{Prec.}
& \textbf{F1}
& \textbf{Recall}
& \textbf{Acc.} \\
\midrule
CWT -- resizing
    & 8.94 & 3.42 & 6.18
    & 78.82 & 85.61 & 95.19 & 87.91 \\
STFT -- resizing
    & 9.85 & 3.62 & 6.73
    & 77.28 & 83.31 & 93.73 & 85.96 \\
Padding -- STFT
    & 9.67 & 4.07 & 6.87 &
    69.75 & 74.88 & 79.97 & 84.78 \\
Padding -- LWPT (Length-aware)
    & 9.08 & 3.75 & 6.41&
    81.05 & 87.39 & 97.43 & 91.78 \\
Padding -- LWPT (ours)
    & 7.94 & 3.17 & 5.56 &
    82.60 & 89.80 & 98.37 & 93.29 \\
\bottomrule
\end{tabularx}

\end{table}

\rev{Table~\ref{tab:abl_LWPT} analyzes the performance of different LWPT configurations with and without the learnable denoising module. The proposed db4-based LWPT with denoising achieves the best performance, yielding the lowest overall MAPE of 5.56\% in Case Study~I and the highest accuracy of 93.23\% in Case Study~II. Removing the denoising module degrades performance, with the db4 wavelet reaching an overall MAPE of 5.90\% and an accuracy of 88.19\%. Haar-based configurations show consistently higher estimation errors and lower classification accuracy, with the non-denoising Haar variant exhibiting an overall MAPE of 6.73\% and an accuracy of 87.91\%.}

\begin{table}[H]
\caption{Impact of the learnable signal denoising module on LWPT performance.}
\label{tab:abl_LWPT}
\renewcommand{\arraystretch}{1.25}
\fontsize{8.5}{10.5}\selectfont
\centering

\begin{tabularx}{\linewidth}{
    p{2.8cm}
    *{7}{>{\centering\arraybackslash}X}
}
\toprule
\multirow{2}{*}{\textbf{Model Type}} &
\multicolumn{3}{c}{\textbf{Case Study 1 (MAPE \%)}} &
\multicolumn{4}{c}{\textbf{Case Study 2 (Classification \%)}} \\
\cmidrule(lr){2-4} \cmidrule(lr){5-8}
& \textbf{$k_\mathrm{p}$}
& \textbf{$k_\mathrm{b}$}
& \textbf{Overall}
& \textbf{Prec.}
& \textbf{F1}
& \textbf{Recall}
& \textbf{Acc.} \\
\midrule
Haar                & 9.68 & 3.78 & 6.73 & 74.19 & 80.19 & 88.47 & 87.91 \\
db4                 & 8.41 & 3.39 & 5.90 & 74.48 & 82.41 & 92.24 & 88.19 \\
Haar + Denoising    & 8.89 & 3.43 & 6.16 & 76.78 & 83.66 & 91.89 & 89.23 \\
db4 + Denoising (ours)
                    & 7.94
                    & 3.17
                    & 5.56
                    & 82.60
                    & 89.80
                    & 98.37
                    & 93.29 \\
\bottomrule
\end{tabularx}

\end{table}

\rev{Table~\ref{tab:FE} compares the proposed framework with existing baseline models across both case studies. For a fair comparison with baseline feature extractors, the WaveletInception–BiGRU model without speed fusion is considered. This model achieves the best overall performance among all compared methods, with an overall MAPE of 6.09\% in Case Study~I and an accuracy of 91.53\% in Case Study~II. The WaveletInception model, corresponding to our variant without recurrent temporal modeling, follows with an overall MAPE of 6.46\% and an accuracy of 90.32\%. Among other deep learning baselines, GoogLeNet~\cite{hajializadehDeep2023} exhibits the strongest performance, achieving an overall MAPE of 7.47\% and an accuracy of 88.93\%, followed by the Laplace–Inception model~\cite{lamprea-pinedaRailwayTrackReceptance2024} with an overall MAPE of 8.72\% and an accuracy of 86.49\%. LSTM-based~\cite{riahisamaniBidirectionalLongShortTerm2025} and VGG~\cite{lockeUsing2020} models rank lower, respectively, with overall MAPEs of 10.56\% and 13.08\% and accuracies of 81.03\% and 78.77\%, respectively.}

\begin{table}[H]
\caption{Performance comparison of the proposed framework against state-of-the-art baseline models.}
\label{tab:FE}
\renewcommand{\arraystretch}{1.25}
\fontsize{8.5}{10.5}\selectfont
\centering

\begin{tabularx}{\linewidth}{%
    p{2.9cm}
    *{7}{>{\centering\arraybackslash}X}
}
\toprule
\multirow{2}{*}{\textbf{Model Types}} &
\multicolumn{3}{c}{\textbf{Case Study 1 (MAPE \%)}} &
\multicolumn{4}{c}{\textbf{Case Study 2 (Classification \%)}} \\
\cmidrule(lr){2-4} \cmidrule(lr){5-8}
& {$k_\mathrm{p}$}
& {$k_\mathrm{b}$}
& {Overall}
& {Prec.}
& {F1}
& {Recall}
& {Acc.} \\
\midrule
VGG \cite{lockeUsing2020}
    & 18.88 & 7.28 & 13.08 & 62.82 & 69.27 & 78.19 & 78.77 \\
LSTM Network \cite{riahisamaniBidirectionalLongShortTerm2025}
    & 15.10 & 6.02 & 10.56 & 67.19 & 72.06 & 80.29 & 81.03 \\
Laplace-Inception \cite{liWaveletKernelNet2022}
    & 12.33 & 5.11 & 8.72 & 72.38 & 79.79 & 88.88 & 86.49 \\
GoogLeNet \cite{hajializadehDeep2023}
    & 11.17 & 3.76 & 7.47 & 77.82 & 82.67 & 92.17 & 88.93 \\
WaveletInception 
    & 9.07 & 3.84 & 6.46 & 76.89 & 86.93 & 97.41 & 90.32 \\
WaveletInception-BiGRU (ours, without speed fusion)
    & 8.64 & 3.54 & 6.09 & 81.22 & 88.56 & 98.02 & 91.53 \\
\bottomrule
\end{tabularx}

\end{table}

\rev{Table~\ref{tab:FE_compu} presents the computational efficiency of the proposed framework compared with baseline models. The WaveletInception–BiGRU model achieves a balanced trade-off between model size and inference speed. It requires only 1.01~M and 1.20~M parameters for Case Study~I and Case Study~II, respectively, making it substantially lighter than GoogLeNet \cite{hajializadehDeep2023}, which has a comparable accuracy. In terms of inference time per sample, the WaveletInception-BiGRU model ranks in the middle, slightly slower than the WaveletInception and Laplace–Inception \cite{liWaveletKernelNet2022} models. However, WaveletInception-BiGRU, with 0.99 and 3.14 milliseconds, respectively, in Case Study I and II, remains significantly faster than GoogLeNet \cite{hajializadehDeep2023} and LSTM Network \cite{riahisamaniBidirectionalLongShortTerm2025} models.}

\begin{table}[H]
\caption{Comparison of computational efficiency and model size between the proposed framework and baseline models from the literature.}
\label{tab:FE_compu}
\renewcommand{\arraystretch}{1.25}
\fontsize{8.5}{10.5}\selectfont
\centering

\begin{tabularx}{\linewidth}{
    p{2.9cm}
    *{4}{>{\centering\arraybackslash}X}
}
\toprule
\multirow{2}{*}{\textbf{Model Types}} &
\multicolumn{2}{c}{\textbf{Case Study 1 }} &
\multicolumn{2}{c}{\textbf{Case Study 2 }} \\
\cmidrule(lr){2-3} \cmidrule(lr){4-5}
& Time (ms)
& Model Size (M)
& Time (ms)
& Model Size (M) \\
\midrule
VGG \cite{lockeUsing2020}
    & 3.09 & 2.23 & 4.69  & 2.69 \\
LSTM Network \cite{riahisamaniBidirectionalLongShortTerm2025}
    & 6.69 & 2.88 & 7.70 & 3.32 \\
Laplace-Inception \cite{liWaveletKernelNet2022}
    & 2.16 & 2.1 & 2.46 & 2.29 \\
GoogLeNet \cite{hajializadehDeep2023}
    & 2.93 & 6.45 & 8.27 & 5.61 \\
WaveletInception
    & 0.93 & 1.25 & 2.24 & 1.65 \\
WaveletInception-BiGRU (ours, without speed fusion)
    & 0.99 & 1.01 & 3.14 & 1.2 \\
\bottomrule
\end{tabularx}

\end{table}

\rev{To incorporate operational conditions such as measurement speed, two mid-level fusion strategies, early and late, are evaluated. In early fusion, the encoded speed tensor is concatenated with vibration signal features before processing by the 1D Inception-ResNet modules for deep feature extraction. In late fusion, this concatenation occurs after high-level vibration features are extracted, as designed in our proposed architecture. Table~\ref{tab:abl_fusion} compares six fusion designs, including CNN-, BiLSTM-, and BiGRU-based speed fusion modules. The proposed late BiGRU-based fusion achieves the best overall performance, with the lowest overall MAPE of 5.56\% in Case Study~I and the highest accuracy of 93.29\% in Case Study~II. The late BiLSTM-based fusion is the second-best configuration, achieving an overall MAPE of 5.81\% and an accuracy of 91.52\%. These results indicate that late fusion consistently outperforms early fusion, while BiGRU-based designs provide additional performance gains. In contrast, CNN-based fusion approaches exhibit the lowest performance across both fusion stages, with overall MAPE exceeding 6.49\% and accuracy below 85.41\%, highlighting their limited ability to capture temporal dependencies and fuse auxiliary speed information.}

\begin{table}[H]
\caption{Ablation study of six fusion modules for incorporating measurement speed as auxiliary information in the proposed WaveletInception-BiGRU model.}
\label{tab:abl_fusion}
\renewcommand{\arraystretch}{1.25}
\fontsize{8.5}{10.5}\selectfont
\centering

\begin{tabularx}{\linewidth}{
    p{2.5cm}
    *{7}{>{\centering\arraybackslash}X}
}
\toprule
\multirow{2}{*}{\textbf{Model Types}} &
\multicolumn{3}{c}{\textbf{Case Study 1 (MAPE \%)}} &
\multicolumn{4}{c}{\textbf{Case Study 2 (Classification)}} \\
\cmidrule(lr){2-4}
\cmidrule(lr){5-8}
& $k_\mathrm{p}$
& $k_\mathrm{b}$
& Overall
& Prec.
& F1
& Recall
& Acc. \\
\midrule
Early CNN-based    & 9.63 & 3.86 & 6.74 & 70.86 & 76.02 & 82.57 & 83.84\\

Early BiLSTM-based   & 8.91 & 3.69 & 6.30 & 76.03 & 82.87 & 93.44 & 87.67 \\

Early BiGRU-based   & 8.67 & 3.63 & 6.15 & 77.55 & 85.29 & 95.69 & 89.18  \\
Late CNN-based     & 9.24 & 3.74 & 6.49 & 71.67 & 77.75 & 84.95 & 85.41 \\
Late BiLSTM-based    & 8.28 & 3.35 & 5.81 & 78.96 & 87.61 & 98.09 & 91.52  \\
Late BiGRU-based (ours)     & 7.94 & 3.17 & 5.56 & 82.60 & 89.80 & 98.37 & 93.29 \\
\bottomrule
\end{tabularx}

\end{table}

\rev{We further evaluate the performance of LSTM- and GRU-based modules for capturing temporal dependencies in our model architecture. Table~\ref{tab:abl_temporal} compares LSTM, GRU, BiLSTM, and BiGRU networks for temporal feature extraction. The BiGRU model achieves the best overall results, with the lowest MAPE of 5.56\% in Case Study~I and the highest accuracy of 93.29\% in Case Study~II. BiLSTM follows, with an overall MAPE of 5.81\% and an accuracy of 91.52\%, outperforming its unidirectional counterpart. Among unidirectional models, GRU outperforms LSTM, with a MAPE of 6.31\% versus 6.73\% and an accuracy of 87.88\% versus 85.61\%. Bidirectional variants introduce only a modest increase in the number of parameters while consistently improving performance over unidirectional architectures.}

\begin{table}[H]
\caption{Ablation study comparing unidirectional and bidirectional GRU and LSTM networks for temporal feature extraction in the proposed WaveletInception-BiGRU model.}
\label{tab:abl_temporal}
\renewcommand{\arraystretch}{1.25}
\fontsize{8.5}{10.5}\selectfont
\centering

\begin{tabularx}{\linewidth}{%
    p{2.5cm}
    *{9}{>{\centering\arraybackslash}X}
}
\toprule
\multirow{2}{*}{\textbf{Model Types}} &
\multicolumn{4}{c}{\textbf{Case Study 1 (MAPE \%)}} &
\multicolumn{5}{c}{\textbf{Case Study 2 (Classification \%)}} \\
\cmidrule(lr){2-5} \cmidrule(lr){6-10}
& {$k_\mathrm{p}$}
& {$k_\mathrm{b}$}
& {Overall}
& {Size (M)}
& {Prec.}
& {F1}
& {Recall}
& {Acc.}
& {Size (M)} \\
\midrule
LSTM
    & 9.38 & 4.08 & 6.73 & 1.11
    & 68.87 & 76.84 & 90.29 & 85.61 & 1.30 \\
BiLSTM
     & 8.28 & 3.35 & 5.81 & 1.22
    & 78.96 & 87.61 & 98.09 & 91.52 & 1.57 \\
GRU
    & 8.97 & 3.66 & 6.31 & 1.02
    & 73.13 & 82.34 & 94.21 & 87.88 & 1.26 \\
BiGRU (ours)
    & 7.94 & 3.17 & 5.56 & 1.03
    & 82.60 & 89.80 & 98.37 & 93.29 & 1.41 \\
\bottomrule
\end{tabularx}

\end{table}

\rev{We evaluate the proposed WI-BiGRU model for estimating track stiffness variations under four scenarios in Case Study~I. Table~\ref{tab:results_case1} summarizes the RMSE and MAPE for both \( k_{\mathrm{p}} \) and \( k_{\mathrm{b}} \). The WI-BiGRU model achieves the best results in the uniform stiffness scenario, with MAPEs of 5.11 for \( k_{\mathrm{p}} \) and 1.86\% for \( k_{\mathrm{b}} \). The most challenging case is the single-sleeper stiffness drop, which results in the highest errors (overall MAPE of 7.81\%). The three-sleeper drop scenario yields moderate performance (overall MAPE of 6.19\%), while transition-zone changes are estimated more accurately (overall MAPE of 4.56\%). Across all scenarios, the model finds \( k_{\mathrm{p}} \) harder to estimate than \( k_{\mathrm{b}} \), with overall MAPEs of 7.94\% and 3.17\%, respectively.}

\begin{table}[H]
\caption{WI-BiGRU estimation accuracy performance across four track stiffness scenarios (Case Study I).}
\label{tab:results_case1}
\renewcommand{\arraystretch}{1.25}
\fontsize{8.5}{10.5}\selectfont
\centering

\begin{tabularx}{\linewidth}{%
    p{3.8cm}
    *{4}{>{\centering\arraybackslash}X}
    p{1.4cm}
}
\toprule
\multirow{2}{*}{\textbf{Case Study I Scenarios}}
& \multicolumn{2}{c}{\textbf{$k_\mathrm{p}$}} 
& \multicolumn{2}{c}{\textbf{$k_\mathrm{b}$}}
& \multirow{2}{*}{\shortstack{\textbf{Overall}\\\textbf{MAPE (\%)}}} \\
\cmidrule(lr){2-3} \cmidrule(lr){4-5}
& RMSE (MN/m) & MAPE (\%)
& RMSE (MN/m) & MAPE (\%) & \\
\midrule
(I) Uniform                  & 15.29 & 5.11 & 0.65 & 1.86 & 3.36 \\
(II) Drops in 1 sleeper      & 21.99 & 10.76 & 1.81 & 4.66 & 7.81 \\
(III) Drops in 3 sleepers   & 19.34 & 8.98 & 1.22 & 3.41 & 6.19 \\
(IV) Transition zone changes& 17.01 & 6.62 & 0.86 & 2.54 & 4.56 \\
All scenarios                & 18.21 & 7.94 & 1.13 & 3.17 & 5.56 \\
\bottomrule
\end{tabularx}

\end{table}

\begin{figure}[H]
    \includegraphics[trim=0cm 1cm 0cm 0cm, clip, scale=0.66]{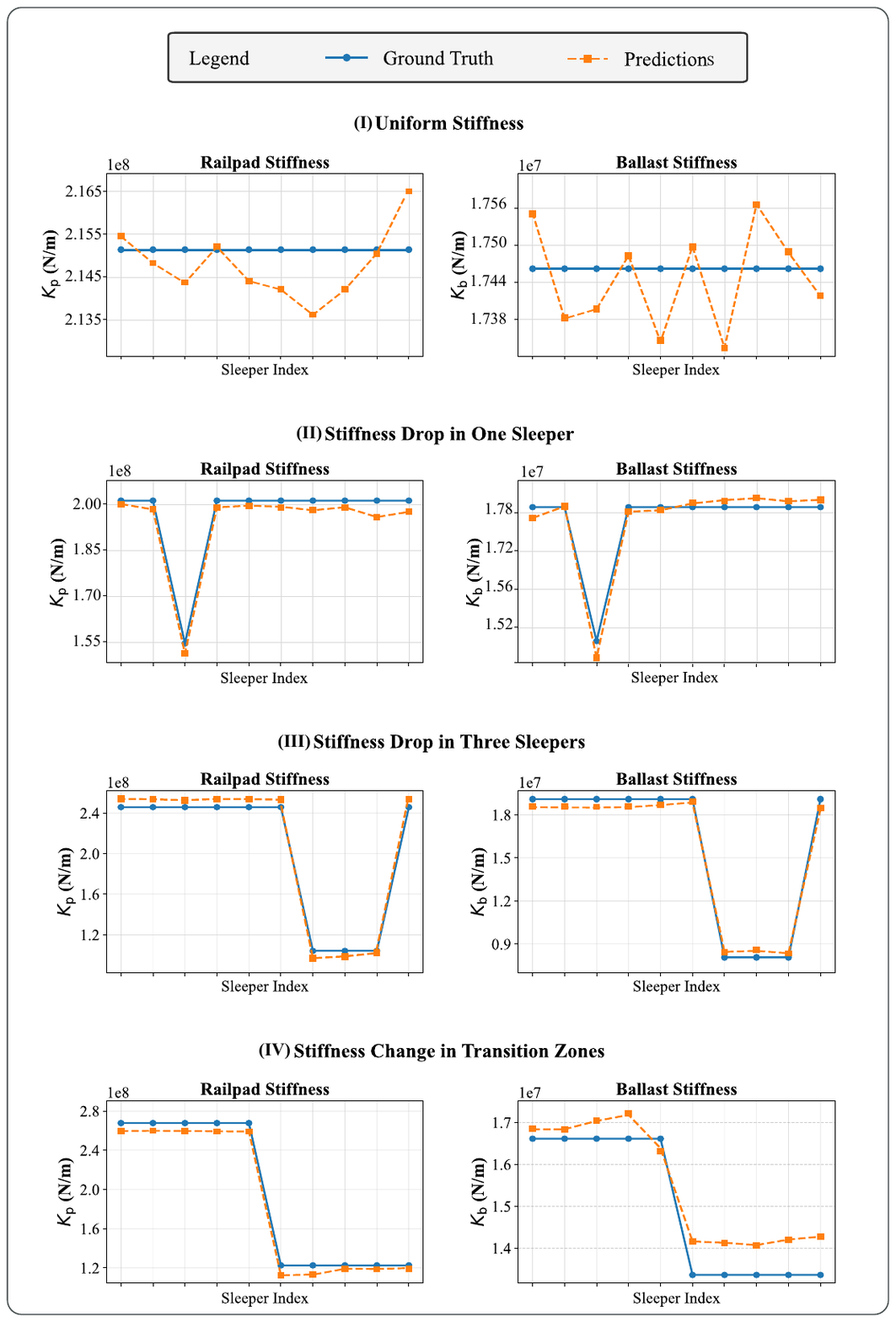}     
      \caption{WI-BiGRU model predictions on the Case Study I test set. Ground-truth values are shown in blue, and model estimates in orange.}
      \label{predictions}
   \end{figure}

Figure \ref{predictions} shows examples of railpad and ballast stiffness estimations compared to their ground truth values. The blue lines represent the ground truth, while the orange lines represent the model’s estimations. These examples illustrate the model's capability to estimate both parameters under various scenarios, including uniform stiffness, localized reductions, and transition zones.

\rev{Furthermore, we evaluate the WI-BiGRU performance for the transition zone identification in Case Study II. Table~\ref{tab:results_case2} summarizes the model's performance in identifying sleepers on the normal track and above the viaduct at Site~4 in Case Study~II. The model achieves an overall accuracy of 93.29\% in both scenarios, demonstrating high and balanced classification performance. For the viaduct stiffness scenario, the model achieves 82.60\% precision, 98.37\% recall, and 89.80\% F1-score, indicating strong detection of transition-zone sleepers with minimal false negatives. In the normal track scenario, the model achieves a precision of 99.24\%, a recall of 91.12\%, and an F1-score of 95.00\%, reflecting reliable identification with few false positives.}

\begin{table}[ht]
\centering
\caption{WaveletInception-BiGRU model estimation accuracy performance on the track transition zone identification (Case Study II).}
\label{tab:results_case2}
\renewcommand{\arraystretch}{1.3}
\fontsize{9}{11}\selectfont
\begin{tabularx}{\linewidth}{l *{4}{>{\centering\arraybackslash}X}}
\toprule
\textbf{Scenarios} & \textbf{Prec. (\%)} & \textbf{Recall (\%)} & \textbf{F1 (\%)} &\textbf{Accuracy (\%)}  \\
\midrule
(I) Normal Track Stiffness & 99.24 & 91.12 & 95.00 & 93.29  \\
(II) Viaduct Stiffness & 82.60 & 98.37 & 89.80 & 93.29  \\
\bottomrule
\end{tabularx}
\end{table}

\section{\rev{Discussion}}\label{Disc}

\rev{This paper presents two case studies demonstrating the effectiveness of the WaveletInception-BiGRU deep learning framework for on-board vibration response analysis. Our model outperforms four external baseline models and its own ablations in both track stiffness estimation and railway transition zone identification. In the first case study, the WaveletInception-BiGRU model achieves a MAPE of 7.94\% for railpad stiffness and 3.17\% for ballast stiffness across different scenarios of track stiffness reduction (see Table~\ref{tab:results_case1}). In the second case study, it yields an accuracy of 93.29\% for transition zone identification (see Table~\ref{tab:results_case2}). Compared to baseline models in the literature, our framework demonstrates notable accuracy (see Table~\ref{tab:FE}) and computational efficiency (see Table~\ref{tab:FE_compu}), supported by insights from our ablation studies (see Tables~\ref{tab:abl_preprocessing}-~\ref{tab:abl_LWPT} and Tables~\ref{tab:abl_fusion}-~\ref{tab:abl_temporal}).}

At the core of the proposed architecture is the WaveletInception network, which integrates an LWPT stem with 1D Inception-ResNet blocks. The LWPT stem decomposes input vibration signals into multi-resolution components while preserving all original signal information due to its orthogonality. This decomposition provides physically informative features in the early stages of the network and enables downsampling without information loss.

\rev{The results in Table~\ref{tab:abl_preprocessing} demonstrate that the proposed LWPT-based stem module significantly outperforms conventional preprocessing techniques. Standard resizing techniques, such as those used for CWT and STFT, rely on interpolation and averaging, which introduce non-physical values and spectro-temporal decimation. This resolution loss is particularly detrimental in rail health monitoring, where accurately localized, transient signatures—such as rail squats or fastening failures—are obscured by the smoothing effects of frequency-bin averaging and reduced temporal resolution.}

\rev{Furthermore, the LWPT stem addresses the critical challenge of handling variable signal lengths via zero padding. While essential for CNN-based deep learning architectures, padding often introduces artificial discontinuities and spectral leakage, leading to the degraded performance observed in STFT-based preprocessing (Accuracy: 84.78\%; MAPE: 6.87\%). In contrast, the proposed LWPT stem mitigates these artifacts through multi-resolution analysis, adaptively resolving frequency components without the constraints of rigid windowing. Specifically, utilizing $db4$ filters over the Haar variant proves superior; the higher-order $db4$ wavelets provide smoother boundary transitions, effectively distributing padding-induced distortions and allowing the model to isolate the low-frequency features essential for accurate defect mapping.}

\rev{Empirical evidence (Table~\ref{tab:abl_LWPT}) also highlights the effectiveness of the employed learnable denoising function \cite{michauFullyLearnableDeep2022}. Given that stochastic noise is inherent in real-world vibration data, these results suggest that integrating adaptive thresholding directly into the architectural stem is more effective than static preprocessing (no denoising).}



\rev{Since not all frequency decomposition bins of the LWPT are equally relevant for the subsequent deeper layer, our analysis showed the significance of a channel projection layer to adapt the LWPT for a supervised deep learning architecture effectively. This can boost the architecture to prioritize frequency decompositions, which are more informative for the task-specific monitoring. However, a primary limitation of the current LWPT implementation is its sequential recursive structure, which precludes parallel computation across nodes. As the decomposition depth $L$ increases, the exponential growth of the number of nodes ($2^L$) can significantly increase computational overhead. Furthermore, a physical interpretation of our stem module, particularly the LWPT and its channel projection layer, remains an open question for future research.}

\rev{The WaveletInception network offers both representational and computational benefits for vibration-based feature extraction. It integrates spectral information directly into the deep learning architecture. The 1D Inception–Residual network has multi-scale parallel branches that capture informative patterns across temporal scales, essential for track condition monitoring \cite{shenEvaluatingRailwayTrack2023a, lamprea-pinedaRailwayTrackReceptance2024, shenFastRobustIdentification2021}. The Inception-ResNet blocks use filters of varying sizes to extract multi-scale, high-level features from the outputs of the stem module.} 

\rev{Moreover, the residual connections support stable learning and an efficient gradient flow, helping deeper models converge and avoid degradation. Bottleneck convolutions further reduce parameter count and computational cost while preserving discriminative power.}

\rev{The proposed 1D convolutional network avoids the computational overhead of 2D convolutions over time–frequency representations (see Table \ref{tab:FE_compu}). This design not only reduces processing time but also can preserve the physical interpretability of the extracted features, as they remain closely linked to the original signal. While 2D networks \cite{hajializadehDeep2023} can capture complex time–frequency interactions, their features are often more abstract and computationally demanding. Indeed, they convolve over frequency bins, which may not be physically interpretable. By focusing on 1D convolutions, the WaveletInception-BiGRU model can achieve an efficient balance between performance, interpretability, and computational cost.}

\rev{The proposed framework leverages Bi-directional Gated Recurrent Units (BiGRU) to model the long-range temporal dependencies inherent in on-board vibration responses. As shown in Table~\ref{tab:abl_temporal} and motivated by Figures~\ref{fg:ave_power}-~\ref{case2SFTF}, the inclusion of bidirectional context is critical for capturing interrelated dependencies between measurement speed, excitation levels, and modal characteristics. By processing the signal in both forward and backward paths, the architecture captures the neighboring context of vibrations, which is essential for the accurate health-condition estimation of individual infrastructure components.}

\rev{The proposed BiGRU-based architecture captures the relationships between onboard vibrations and operational-condition embeddings. As shown in Table 7, the late-fusion BiGRU strategy significantly outperforms CNN and LSTM counterparts, particularly in the real-world scenarios of Case Study II (93.29\% accuracy). This advantage arises from the BiGRU’s ability to effectively model interactions between speed embeddings and spectral features extracted by the WaveletInception network. By capturing bidirectional temporal dependencies, the network adaptively manages variable signal lengths, mitigates zero-padding artifacts, and adjusts the extracted vibration features under operational conditions, ensuring consistent health-condition estimation. Furthermore, the high precision observed in Case Study II, which includes environmental noise and sensor interference typical of real-world conditions, demonstrates the framework's practical robustness. While this paper confirms the model’s reliability across diverse operational conditions, future work should systematically stress-test the model under extreme signal-to-noise ratios to further assess the noise robustness of our proposed model architecture.}

\rev{A distinctive feature of our architecture is the distributed temporal-sampling strategy that generates $N$ feature vectors for the corresponding $N$ structural beams. Recurrent models, such as LSTM and GRU modules, often employ a many-to-one architecture, collapsing the entire sequence into a single terminal hidden state (the last time step). This can create an information bottleneck, particularly for high-resolution vibration signals, in which local transient features are lost during global compression. In contrast, our strategy performs distributed selection across the temporal dimension, maintaining a direct mapping between features and the physical, spatial distribution of the target component within the infrastructure.}

\rev{This hierarchical sequence modeling approach offers significant practical advantages for on-board monitoring. By selecting $N$ equidistant points in the temporal domain, the model effectively aligns with the physical layout of the infrastructure without requiring high-precision spatial positioning. This is particularly advantageous when measurement speed varies; while precise GPS-based positioning becomes challenging under fluctuating speeds, the number of structural targets within a segment remains constant. Consequently, the distributed sampling preserves temporally localized features and global context, enabling finer localized health assessments while mitigating the representation loss inherent in standard length-aware or packing-based sequence modules (see Table~\ref{tab:abl_preprocessing}).}

\rev{By processing vibration responses directly in the time domain, the proposed architecture eliminates the necessity for manual feature engineering. Unlike traditional diagnostic frameworks that rely on hand-crafted statistical or spectral indicators, the proposed stem module enables the network to automatically derive sensitive features directly from vibration signals collected at varying speeds. This process is further enhanced by the BiGRU-based integration of operational data, which facilitates the internal modeling of vibration signatures across varying speed profiles. Consequently, this end-to-end approach establishes a fully automated monitoring system that maintains high diagnostic precision despite fluctuations in measurement speed.}

\rev{
The WI-BiGRU model demonstrates high precision across all scenarios in Case Study I (Table~\ref{tab:results_case1}), though performance varies with the localization of defects. While the overall MAPE is 5.56\%, the highest misestimation occurs in Scenario (II) (single sleeper drop), where the $k_\mathrm{p}$ error reaches 10.76\%. This indicates that highly localized, transient stiffness changes are more difficult to isolate than the sustained signatures found in 3-sleeper drops or transition zones. Furthermore, rail pad stiffness ($k_\mathrm{p}$) exhibits higher error rates than ballast stiffness ($k_\mathrm{b}$), likely because $k_\mathrm{b}$ has a more dominant influence on the global track vibration response. In Case Study II (Table~\ref{tab:results_case2}), the model accurately identifies transition zones but reveals a lower precision for the viaduct stiffness class (82.60\%). This suggests that some normal sleepers near the transition boundary are misclassified as viaduct sleepers due to mechanical coupling and boundary blurring across different track structures. These findings indicate that, although the model is highly effective for mapping structural conditions, its sensitivity can still be improved for single-point defects and boundary regions.
}

\rev{The dataset in Case Study I is based on a vehicle–track interaction model from \cite{shenEvaluatingRailwayTrack2023a} to demonstrate our deep learning framework. Simulated drive-by responses may not capture all real-world environmental and operational factors. Phenomena like weld effects, temperature changes in rail stiffness, and rail neutral temperature are not modeled. The simulation uses a single vehicle–track setup, while real railways have varying profiles, welds, fastenings, sleepers, and structures that could affect results \cite{shenEvaluatingRailwayTrack2023a}. Case Study II shows that our framework can identify transition zones in real-world conditions. Transition zones may exhibit additional degradation mechanisms, such as ballast fouling and differential settlement, suggesting that expanding the model to include additional health indicators, such as track geometry degradation and differential settlement, is essential for future research.  Additionally, the framework can also be applied to various track types, including ballast and ballastless tracks, and broader transportation infrastructure, such as railway and road bridges and tunnels.}


\section{Conclusions and future work}\label{Conc}

In this paper, we have proposed a novel deep learning framework, Wavelet-Inception-BiGRU, for estimating infrastructure health condition monitoring using onboard vibration signals. The core of our methodology is a new feature-extraction network, called WaveletInception, that integrates a Learnable Wavelet Packet Transform (LWPT) and 1D Inception-ResNet blocks to effectively extract multi-scale, high-level representations from vibration signals. Furthermore, the feature extraction network incorporates operational conditions, in particular measurement speed, through feature-level fusion with a Long Short-Term Memory (LSTM) layer. This enables the model to learn interrelated features across different measurement conditions. \rev{For the health condition estimation head, we have proposed a sequential model using BiGRU networks.} The model leverages temporal information from both the forward and backward directions of on-board measurements. Additionally, the sequential modeling approach enables more detailed estimation at the beam or component level, enabling comprehensive assessments of different segments of infrastructure.


\rev{Future research will focus on three main directions. First, this paper is one of the first to include operational conditions of on-board vibration response in a deep learning model. The BiGRU layer captures temporal dependencies and fuses vibration and operational data. In the future, other temporal encoders, like attention mechanisms, can be explored to improve feature extraction. Second, our results show that the proposed model works well for health indicators with low-frequency features. Next, the framework can be adapted for high-frequency applications, such as defect detection in superstructures and rails, using advanced sensors like laser Doppler vibrometers. Third, to support large-scale, automated, cloud-based monitoring, future research will systematically test the framework’s stability when signal-to-noise ratios and operational conditions vary.}

\medskip\section*{Acknowledgments}

The first author thanks Hesam Araghi for insightful comments and Chen Shen for providing the simulator for Case Study I. This research was partly supported by ProRail and Europe’s Rail Flagship Project IAM4RAIL - Holistic and Integrated Asset Management for Europe’s Rail System [grant agreement 101101966].

\section*{Disclaimer}

Funded by the European Union. Views and opinion expressed are however those of the author(s) only and do not necessarily reflect those of the European Union or Europe’s Rail Joint Undertaking. Neither the European Union nor the granting authority can be held responsible for them. The project FP3-IAM4Rail is supported by the Europe's Rail Joint Undertaking and its members.


\newpage

\bibliographystyle{elsarticle-num}
\bibliography{References/ITSC2024v9} 

\end{document}